\pdfoutput=1

\documentclass[11pt]{article}

\usepackage{acl}

\usepackage{times}
\usepackage{latexsym}
\usepackage{graphicx}
\usepackage[T1]{fontenc}

\usepackage[utf8]{inputenc}

\usepackage{microtype}
\usepackage[ruled, vlined, noend]{algorithm2e}
\usepackage{amsmath,amssymb,amsfonts,amsthm}
\usepackage{multicol,multirow,booktabs}

\usepackage{inconsolata}

\DeclareMathOperator*{\argmin}{arg\,min}

\newcommand{\eat}[1]{}

\title{\textsc{NeuroPrune}: A Neuro-inspired Topological Sparse Training Algorithm for Large Language Models}

\author{\bf
        Amit Dhurandhar$^{*}$,
        Tejaswini Pedapati$^{*}$,
        Ronny Luss$^{*}$, \\
        \bf
        Soham Dan,
        Aurelie Lozano, Payel Das and Georgios Kollias \\ \\
        IBM Research, Yorktown Heights, NY} 

\begin{document}
\maketitle
\begingroup\def\thefootnote{*}\footnotetext{Equal contribution}\endgroup

\begin{abstract}
Transformer-based Language Models have become ubiquitous in Natural Language Processing (NLP) due to their impressive performance on various tasks. However, expensive training as well as inference remains a significant impediment to their widespread applicability. While enforcing sparsity at various levels of the model architecture has found promise in addressing scaling and efficiency issues, there remains a disconnect between how sparsity affects network topology. Inspired by brain neuronal networks, we explore sparsity approaches through the lens of network topology. Specifically, we exploit mechanisms seen in biological networks, such as preferential attachment and redundant synapse pruning, and show that principled, model-agnostic sparsity approaches are performant and efficient across diverse NLP tasks, spanning both classification (such as natural language inference) and generation (summarization, machine translation), despite our sole objective not being optimizing performance. \textsc{NeuroPrune} is competitive with (or sometimes superior to) baselines on performance and can be up to $10$x faster in terms of training time for a given level of sparsity, simultaneously exhibiting measurable improvements in inference time in many cases.  
\end{abstract}

\section{Introduction}
In the past decade, transformer-based models \cite{attention2017} leveraging \textit{attention} mechanisms have led to state-of-the-art performance
on NLP tasks and other multimodal applications, in both classification and generation settings. Despite the performance improvements, the computational overhead required for training, and inference, hinders progress. The models are large and are typically parameterized by many dense matrices, which also begs the question as to whether this complexity is necessary for better performance.

Sparsity in general neural networks has been considered using sparse regularizations on weights \cite{seeingisbelieving2024} and weight thresholding/masking \cite{dst_2020}.
Specifically for transformers, various attention masking patterns have been studied \cite{sparsebert}. Another direction for inducing sparsity is to remove entire attention heads altogether \cite{michel_19}. Previous sparse methods, however, give little emphasis to the topology of the networks being trained \cite{Xia2022_CoFI}.

In this paper, we study how certain network topologies can be exploited in transformer-based large language models (LLMs) to offer sparser models (in terms of fewer parameters and fewer attention heads overall) while maintaining performance. Our framework, which we call \textsc{NeuroPrune}, is model-agnostic as well as task-agnostic, and is a dynamic sparse training method inspired by biological neuronal networks present in the brain. For example, \citep{yuan2019structural} discusses two stages by which connections (synapses) in a neuronal network evolve in the brain. First, an overabundance of synapses is created, which is similar to the pretraining an LLM. In the second stage, synapses are judiciously removed until stability in the network is achieved, which is akin to fine-tuning an LLM for a particular task by inducing sparsity, or at a higher level, by removing attention heads. Our framework relates sparsity within the Multi-Layer Perceptron (MLP) layers and the attention heads, as well as sparsity at the level of attention heads, to two distinct processes that take place during that second stage of neuronal network development: \emph{preferential attachment} (i.e. rich-get-richer) \citep{lynn2024heavy} and the \emph{elimination of redundancy} \citep{Wang_11}.

\begin{figure}[t]
    \centering
    \includegraphics[width=\columnwidth]{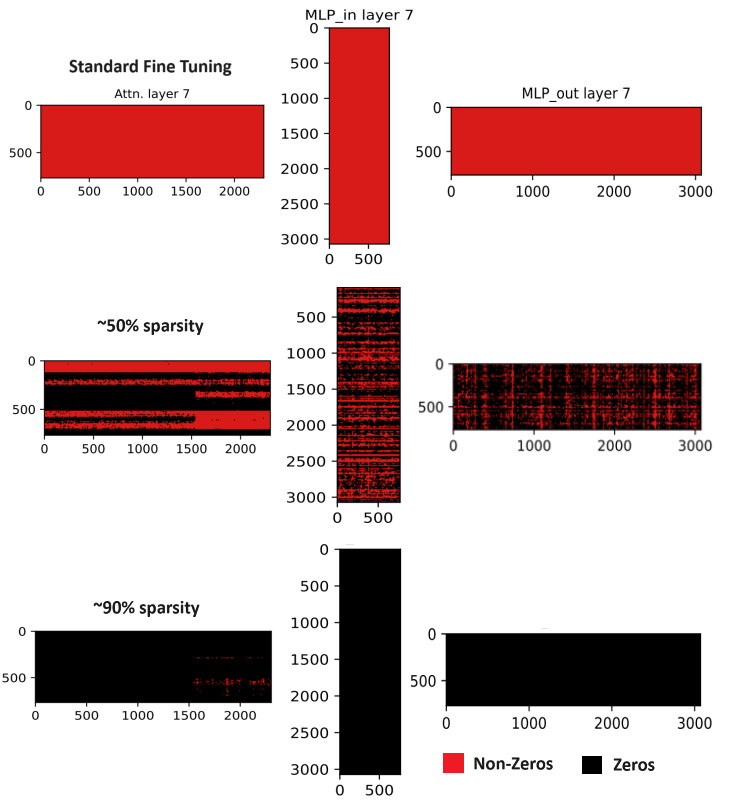}
    \caption{\small{Resulting sparsity patterns ($\approx 50\%$, $\approx 90\%$) as determined by \textsc{NeuroPrune} in an intermediate Transformer layer of a BERT(-base) model learned on the SST2 GLUE benchmark dataset \cite{glue}. \textsc{NeuroPrune} sparsifies according to a preferential attachment topology as entire rows/columns of the attention and MLP matrices are zeroed out. Quantitatively, the standard deviation (sd) between the connectivity of neurons in the MLP layers increases up to two orders of magnitude ($50\%$: $12.12$, $90\%$: $4.16$) compared with standard fine-tuning ($0.13$) as seen in Figure \ref{fig:topo_sparsity_mlp}, This increase in sd is indicative of preferential attachment, similar to what is seen in biological neurons \cite{lynn2024heavy}, where a minimal \textit{rich-get-richer} mechanism is used to produce sparse and heavy-tailed networks. The pattern is qualitatively similar for other layers too, as can be seen in the Appendix.}
    }
    \label{fig:topo_sparsity_intro}
\end{figure}
\begin{figure}[htbp]
    \vspace{-.25cm}
    \centering
    \includegraphics[width=.45\columnwidth]{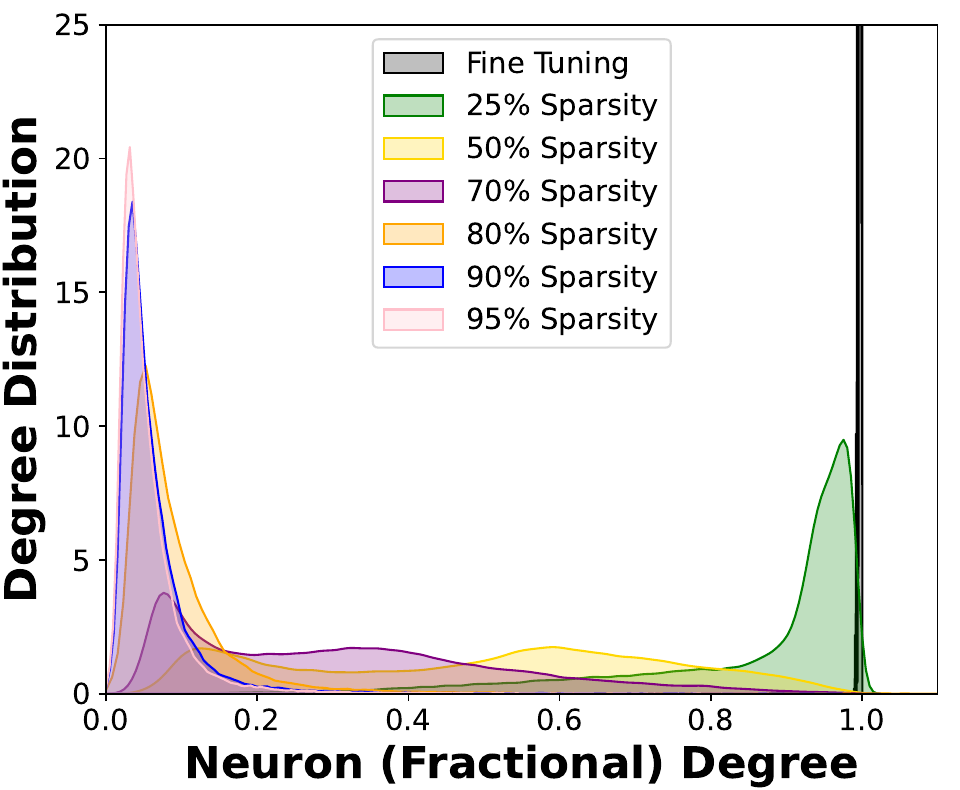}
    \includegraphics[width=.5\columnwidth]{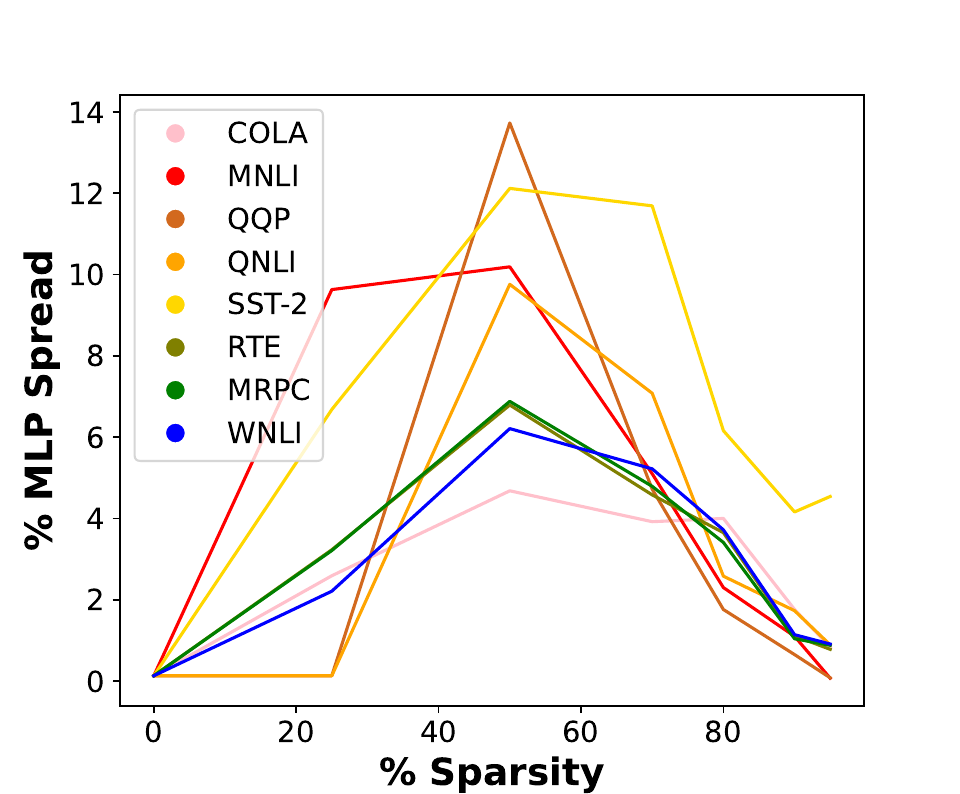}
    \caption{\small{Left is the MLP degree distribution for the SST2 dataset using a BERT model indicative of preferential attachment for \textsc{NeuroPrune} as sparsity increases (echoing the degree distribution in brain functional networks~\cite{vertes2012simple}). Standard fine-tuning creates a dense network (black vertical line). Right we see the non-uniformity in connectivity at different sparsity \%s across GLUE tasks using \textsc{NeuroPrune}, indicative of this preferential attachment across tasks.}
    }
    \label{fig:topo_sparsity_mlp}
\end{figure}

\noindent \textbf{Preferential attachment inspired regularization} Within MLP layers and attention heads, \textsc{NeuroPrune}, is motivated by a well-known network concept, called \textit{preferential attachment}, which was found to be highly relevant in neuronal networks in the brain \cite{vertes2012simple, lynn2024heavy} over the last two decades. The general notion is that over time, neurons with more connections build even more connections, while those with fewer connections are removed. Similarly, our framework induces weighted $l_1$ sparsity in MLP layers (weight inversely $\propto$ connectivity/degree) and group sparsity within attention heads, so that influential neurons (measured by attention parameters) are maintained while those with little influence are pruned. Modeling the removal of weak synapses is an established approach \cite{chechik_1999} to understanding the refinement process of neurons in the brain. For LLMs, this effort is illustrated in Figure \ref{fig:topo_sparsity_intro} where, \textsc{NeuroPrune}, sparsifies the parameter matrices of transformers by zeroing out entire rows in attention and MLP layers. Quantitatively, this non-uniformity in connectivity is verified in Figure \ref{fig:topo_sparsity_mlp}.

\begin{figure}[t]
    \centering
    \includegraphics[width=.49\columnwidth]{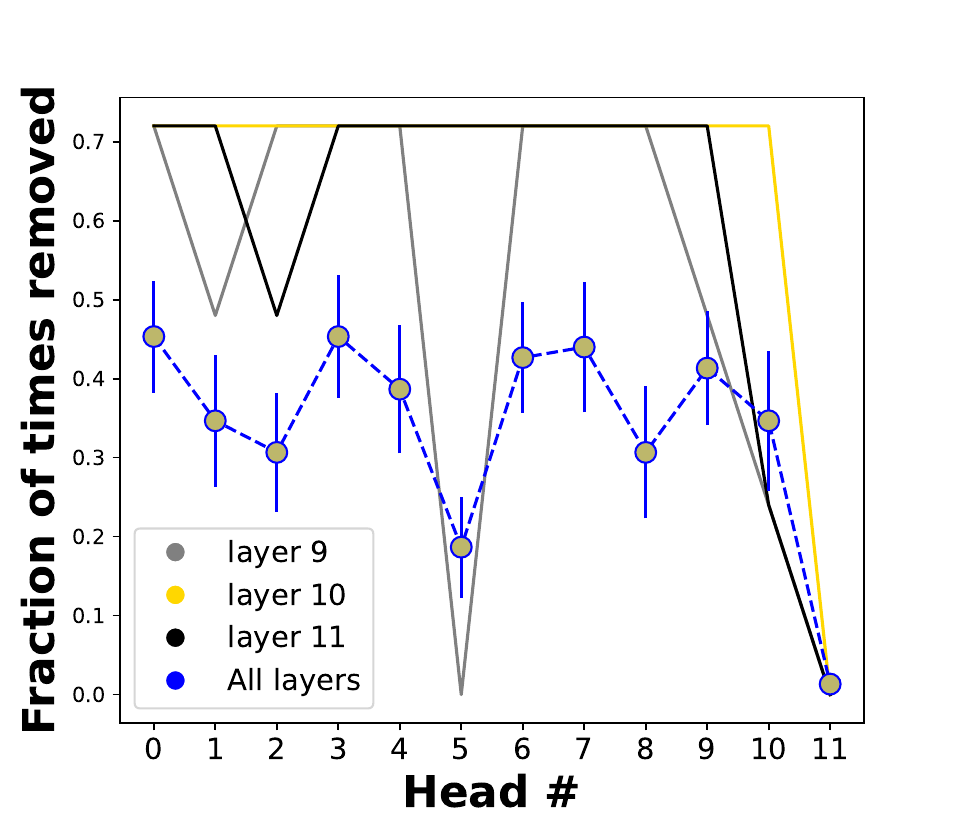}    \includegraphics[width=.49\columnwidth]{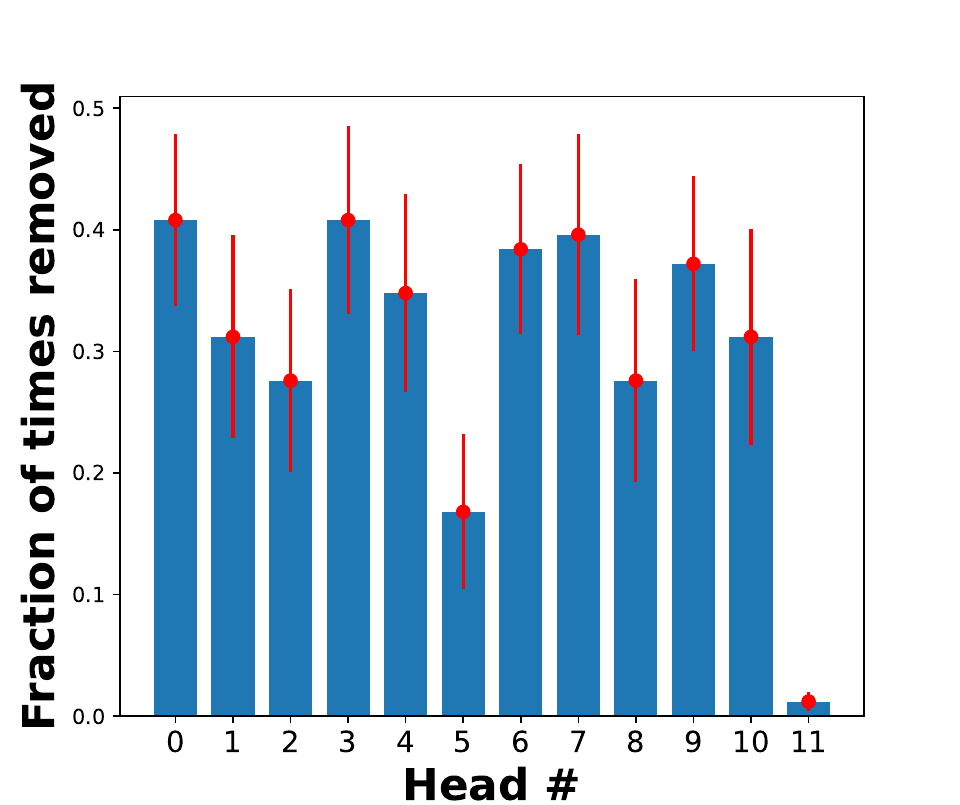}
    \caption{\small{Left is the fraction of times a head is removed using \textsc{NeuroPrune} when fine-tuning on SST2 with a BERT model. The overall numbers (blue curve) are averaged across layers and runs ($\pm$ sd), where at least $10$ heads are removed. We also show individual layer numbers averaged across runs for the top three layers where most pruning of heads happens (Figure \ref{fig:headimp}(right)). We see there is a significant bias towards keeping the last head in each layer leading to a more modular structure and also showcasing preferential attachment, as neurons from the previous and next layers will connect only to these heads. The middle head (head $5$) is also maintained more than most other heads, as it replaces many of the earlier heads. Results averaged across GLUE tasks on the right are similar.}
    }
    \label{fig:topo_head_intro}
\end{figure}

\noindent\textbf{Redundancy-based pruning} While structured sparsity aims at preferential attachment, such pruning (by zeroing out weights) cannot determine which connections are redundant. Elimination of redundant connections is an important aspect of the refinement process \cite{Wang_11, hashimoto_2013} that takes place after the brain develops very dense networks of connections. In the case of LLMs, we hypothesize that such redundancy can be measured by similarity between attention heads, whereupon similar attention heads can then be merged, resulting in reduced complexity while maintaining functionality (i.e. performance is maintained on downstream tasks). Such removal of redundancy is conjectured in \citet{Lichtman_2000} to be unique to the neuron development in the central nervous system of vertebrates. 
Figure \ref{fig:topo_head_intro} illustrates how often heads are found to be redundant using \textsc{NeuroPrune}. Generally, the last head is found to be the least redundant, while the middle head also exhibits limited redundancy. In Figure \ref{fig:headimp}(right), we see head removal as a function of layers and find that the last three layers have the highest number of redundant heads.

\noindent\textbf{Contributions} In this paper, we propose a neuro-inspired topological sparse training algorithm with custom attention and MLP (structured) sparsity regularizations based on preferential attachment, and a novel redundancy-based head pruning scheme, which we map to the dominating set problem~\cite{domset} in theoretical computer science.

Our approach has the following benefits:
\vspace{-\topsep}
\begin{itemize}
 \setlength{\parskip}{0pt}
 \setlength{\itemsep}{0pt plus 1pt}
  \item It is task agnostic.
  \item It is easily adaptable to different transformer-based LLM architectures as it does not add additional (mask) variables to do the pruning. We apply it to BERT (encoder), T5 (encoder-decoder) and OPT (decoder) models. 
  \item It learns sparsity patterns exhibiting principled topological structure. 
  \item It results in LLMs with a competitive and even sometimes superior performance on different benchmarks and tasks (GLUE, summarization, machine translation), although our proposal is more neuroscience-motivated than solely trying to maximize performance.
  \item It is generally much faster to train than the competing baselines, with time per epoch being similar to standard fine-tuning. It also exhibits inference speedups as the topological constraints encourage $N$:$M$-type sparsity.

\end{itemize}
\vspace{-\topsep}

\section{Notation}

A Transformer consists of multiple identical units. Each unit in turn is comprised of a Multi-Head Attention (MHA) Layer and a Feed Forward (FFN) or MLP  Layer (used interchangeably). Each attention layer is partitioned into multiple heads $H_{i}$ composed of Query $Q_{H_{i}}$, Key $K_{H_{i}}$ and Value $V_{H_{i}}$ parameter matrices. If $d$ denotes the embedding dimension of each token in an input matrix $X$ then,
\begin{align*}
     H_{i} = softmax\left(\frac{XQ_{H_{i}} K_{H_{i}}^{T}X^T}{\sqrt{d}}\right) XV_{H_{i}}
 \end{align*}  
A MHA layer with $k$ heads computes the attention of all heads in parallel and concatenates them. $MHA = Concat(H_{1},\ldots, H_{k})W^{O}$, where $W^{O}$ is an output dense matrix.

The FFN layer in turn has two linear layers, one to expand the dimensions $L_{in}:\mathcal{R}^{d_e}\times \mathcal{R}^{d}$ and the other to project it back to the original dimension $L_{out}:\mathcal{R}^{d}\times \mathcal{R}^{d_e}$. Typically $d_e>>d$ (e.g. in BERT $d_e=3072$ and $d=768$) with $L$ denoting the concatenation of all the MLP layers.

If $Q$, $K$, $V$ are the Query, Key, and Value matrices of all the heads in a MHA layer concatenated together, then  $Q_{H_{i}}$, $K_{H_{i}}$ $V_{H_{i}}$ corresponds to the $(i-1)\frac{d}{k}+1$ to $i\frac{d}{k}$ columns in each such matrix respectively. Let $L_{in,\mathcal{H}_i}$ denote the corresponding columns of the MLP and let $A_{\mathcal{H}_i}=[Q_{\mathcal{H}_i}, K_{\mathcal{H}_i}, V_{\mathcal{H}_i}]$. 

The use of the superscript in $A^{(l)}$ denotes (attention) layer $l$ of the transformer. We use $\overrightarrow{\textbf{x}}$ to signify that $x$ is a vector.

\section{Related Work}
Quantization~\cite{ahmadian2023intriguing}, Knowledge Distillation to a smaller model~\cite{gu2023knowledge}, and Model Pruning are some ways to alleviate the extensive computational cost required by LLMs.  
Here we review prior work on model pruning, which is most relevant to us.

\subsection{Unstructured Pruning}
Unstructured pruning \cite{Frantar_2023_SparseGPT, Sun2023_Effective_Pruning, Han2015_Deep_Compression, Tanaka2020_Synflow, Lee2019_SNIP} removes the less salient parameters from the model, thereby achieving sparsity. 
Based on the lottery ticket hypothesis, \citet{Frankle2019_Lottery_Ticket} performs iterative magnitude pruning. 
\citet{Prasanna2020_Bert_Lottery, Chen2020_Lottery_Bert} apply the lottery ticket hypothesis to the BERT model.
This class of pruning algorithms attain high sparsity while largely maintaining accuracy, but are mostly post hoc. Moreover, the resulting pruned models do not provide much inference speedup.

\subsection{Structured Pruning} A neural network can be divided into blocks or components.
For instance, channels, kernels, and layers for a convolution neural network, an attention head, and a fully connected layer for a transformer.
Structured Pruning \cite{Anwar2015_StructuredPruning, Fan2020_Structured_Dropout} involves removing an entire component, thus eliminating some of the multiply-and-accumulate computations, thereby accelerating inference.

\paragraph{Head Pruning}
 \citet{michel_19} were the first to examine if all the heads are necessary for a BERT model. They defined the \textit{importance of a head} by the drop in performance of the model upon removing the head. 
\citet{Viota2019_Head_Pruning} apply gates to each head and learn these gates using $l_0$ regularization. 
\citet{Li2021_DSP} also learn gates and identifies a subset of heads for each layer of the BERT model such that the drop in the model's performance is minimal. They sample the top-k heads based on their importance score and use the Gumbel-softmax trick to make the top-k formulation differentiable. This method was shown to have superior results to other competitors on BERT, and we thus compare it with \textsc{NeuroPrune}'s head pruning strategy.

\paragraph{Block and Layer Pruning} \citet{Fan2020_Structured_Dropout} and \citet{Sajjad2020_Poor_Bert} 
 experimented with dropping different transformer layers, such as every alternate layer, or top-k layers, or middle layers, and found inference speedups. 
\citet{Lagunas2021_Block_Pruning} divide the MHA and FFN layers into several blocks, and apply masks to each of the blocks to prune them. 

CoFI \cite{Xia2022_CoFI} also prunes a transformer by applying gates to each of the heads $m_{h}$, one mask to the entire MHA layer,  and finally, one to the MLP layer in the block.
The model is then trained using $l_0$ regularization to learn these gates. The sparsity constraints are imposed using Lagrangian multipliers. 
To further boost the performance of the pruned model, CoFI jointly prunes and performs layer-wise distillation.  

As \textsc{Neuroprune} also prunes the attention matrices, the feed-forward layer, and attention heads, this is the closest baseline when varying the percentage of sparsity. Note that in contrast to CoFI, we do not require any additional mask variables where an architecture has to be modified, and hence, our approach is easily transferable across model architectures. 
We demonstrate this via experiments on BERT (an encoder-only model), T5 (an encoder-decoder model), and OPT (a decoder-only model).

\section{Method}
We propose (topological) sparsification of a Transformer block at three levels: i) The two (expand and contract) Multi-Layer Perceptron (MLP) layers, ii) the attention layers, and iii) head pruning at the level of attention layers. Our method, \textsc{NeuroPrune}, is detailed in Algorithm \ref{algo:neuroprune} with two sub-procedures given in Algorithms \ref{algo:headprune} and \ref{algo:dominating}. The three sparsifications are described next.

\begin{algorithm}[htbp]
\SetAlgoLined
\textbf{Input:} Model $M(\cdot;A_0,L_0)$ with $N_L$ layers, initial $Q$, $K$, $V$ concatenated matrices $A^{(l)}_0$ and MLP weight matrices $L^{(l)}_0$ for layer $l$, sparsity parameters $\alpha,\beta\ge 0$, \# epochs $n_e$, loss function $\lambda$, redundancy threshold $\theta$

\For{$i=1$ to $n_e$}
{
$
(A_i,L_i)\leftarrow SGD\bigg(\lambda(M(\cdot;A_{i-1},L_{i-1}))+\sum^{N_L}_{l=1}\alpha R_{attn}^{(l)}(A^{(l)}_{i-1})+\beta R_{mlp}^{(l)}(L^{(l)}_{i-1})\bigg) 
$
\# See Eq (\ref{eq:mlp}-\ref{eq:attn}) for $R_{mlp}^{(t)}(\cdot)$, $R_{attn}^{(t)}(\cdot)$

\For{$l=1$ to $N_L$}{
\# See Algorithm \ref{algo:headprune}\\
$(A^{(l)}_i, L^{(l)}_i)\leftarrow \mbox{ELIM\_REDUNDANT}(A^{(l)}_i, L^{(l)}_i,\theta)$
}
}

\KwOut{$(A_{n_e}, L_{n_e})$}

\caption{NeuroPrune}
  \label{algo:neuroprune}
  
\end{algorithm}

\eat{
\begin{algorithm}[htbp]
\SetAlgoLined
\textbf{Input:} Model $M$, $n_b$ data batches $b_1,...,b_{n_b}$, attention and MLP sparsity parameters $\alpha,\beta\ge 0$, \# epochs $n_e$, loss function $\lambda$, distance threshold $\theta$

\For{$i=1$ to $n_e$}
{
\For{$j=1$ to $n_b$}
{
$M\leftarrow \argmin \lambda(M(b_j))+\sum_{\text{layers~} t\in M}\alpha Reg_{attn}^{(t)}+\beta Reg_{mlp}^{(t)}$ 

\# see eqns. \ref{eq:mlp}, \ref{eq:attn}
}

$M\leftarrow$ Call Algorithm \ref{algo:headprune} for each attention layer $L$ of model $M$ with threshold $\theta$ and return model
}

\KwOut{$M$}

\caption{NeuroPrune}
  \label{algo:neuroprune}
  
\end{algorithm}
}

\begin{algorithm}[t]
\SetAlgoLined

\textbf{Input:} Attention layer $A=[K, Q, V]$, MLP layer $L$, $k$ attention head index subsets $~\mathcal{H}_1,~...,~\mathcal{H}_k$, distance threshold $\theta$

\textbf{Initialize:} $S=I_k$~~~\# $k$ dim. identity matrix

\textbf{\# Find similar heads}

$S[i,j] = 1\{l_{\infty}(A_{\mathcal{H}_i}, A_{\mathcal{H}_j}) \leq \theta\}\quad \forall i,j$

$no\_similar = 1\{\sum_{i\neq j}{S[i,j]} > 0\}$

\If{$no\_similar = 0$}
{
\KwOut{(A,L)}
}

\textbf{\# Find redundant heads using similarity}

\# See Algorithm \ref{algo:dominating}\\
$D\leftarrow \mbox{FIND\_DOMINATING}(S)$

\textbf{\# Adjust $(A,L)$ for redundant heads}

\For{$i=1$ to $k$}
{
$\mathcal{J}=\{\mathcal{H}_j|D[i,j]=1, i\neq j\}$

\For{$\textbf{J}\in\mathcal{J}$}
{
$L_{in,\textbf{J}} = L_{in,\textbf{J}}+L_{in,\mathcal{H}_i}$

Prune$(A_{\mathcal{H}_i},L_{\mathcal{H}_i})$ from $(A,L)$ if $\mathcal{J}\neq \phi$

}
}

\KwOut{$(A, L)$}
  \caption{ELIM\_REDUNDANT}
  \label{algo:headprune}
  \end{algorithm}

\eat{
\begin{algorithm}[htbp]
\SetAlgoLined

\textbf{Input:} an attention layer $L$ with $k$ heads $h_1,~...,~h_k$, distance threshold $\theta$

\textbf{Initialize:} $S=I_k$~~~\# $k$ dimensional similarity matrix set to identity

$D=I_k$~~~\# $1$ indicates head in row $i$ can be replaced by head in column $j$

\textbf{\# Find similar heads}

$no\_similar = 0$

\For{$i=1$ to $k$}
{
\For{$j=1$ to $k$}
{
\If{$i\neq j$ and $l_{\infty}(h_i, h_j)\le \theta$}
{
$S[i,j] = 1$

$no\_similar = 1$
}
}
}

\If{$no\_similar = 0$}
{
\KwOut{$L$}
}

\textbf{\# Find redundant heads using similarity}

\For{$i=1$ to $k$}
{
$dom = i$

\For{$j=1$ to $k$}
{
$domvec = S[dom,:]-S[j,:]$

\If{($1\notin domvec$ and $-1 \in domvec$) or ($domvec=\vec{0}$ and $dom < j$)}
{
$dom = j$
}
}
$D[i,dom]=1$
}    

\textbf{\#Adjust for redundant heads and prune}

\For{$i=1$ to $k$}
{
Add output dense weights (and biases) of $h_i$ to output dense weight of $H=\{h_j|D[i,j]=1, i\neq j\}$

Prune $h_i$ from $L$ if $H\neq \phi$
}

\KwOut{$L$}
  \caption{Redundancy-based topologically guided head pruning in NeuroPrune.}
  \label{algo:headprune}
  
\end{algorithm}
}
\begin{algorithm}[t]
\SetAlgoLined

\textbf{Input:} Similarity matrix $S$ for $k$ points

\textbf{Initialize:} $D=I_k$~~~\# $D[i,j]=1$ indicates point $i$ can be replaced by point $j$

\textbf{\# Find redundant points using similarity}

\For{$i=1$ to $k$}
{
$j^* = i$

\For{$j=1$ to $k$}
{
$\overrightarrow{\textbf{s}} = S[j^*,:]-S[j,:]$

\If{($1\notin \overrightarrow{\textbf{s}}$ and $-1 \in \overrightarrow{\textbf{s}}$) or \\\quad\quad($\overrightarrow{\textbf{s}}=\vec{0}$ and $j^* < j$)}
{
$j^* = j$
}
}
$D[i,j^*]=1$
}    

\KwOut{$D$}
  \caption{FIND\_DOMINATING}
  \label{algo:dominating}
  
\end{algorithm}

\subsection{MLP Sparsification}
Preferential sparsification of the MLP layers is conceptually the simplest component of \textsc{NeuroPrune}. For each $L_{in}$ and $L_{out}$ matrix in each Transformer layer, a weighted $l_1$ penalty is added to the training objective, where the weights for each row of entries in the matrix are inversely proportional to the (fractional) connectivity of that neuron. Specifically, let $n_{in,i}$ be the number of entries in the $i^{\text{th}}$ row of $L_{in}$ with absolute values less than some small $\epsilon>0$ (with $n_{out,i}$ similarly defined for $L_{out}$). The MLP regularizer added to the training loss for layer $l$ is as follows:
\begin{align}
\label{eq:mlp}
    R_{mlp}^{(l)}(L^{(l)}) = &\frac{1}{d}[n^{(l)}_{in,1},...,n^{(l)}_{in,d_e}]\cdot|L_{in}^{(l)}|\cdot\vec{1}_d\\&+\frac{1}{d_e}[n^{(l)}_{out,1},...,n^{(l)}_{out,d}]\cdot|L_{out}^{(l)}|\cdot\vec{1}_{d_e} \nonumber
\end{align}
where, $|.|$ denotes an element-wise absolute value and $\vec{1}_d$ is a $d$-dimensional vector of $1$s. In essence, Equation \eqref{eq:mlp} penalizes neurons with less connectivity more than the densely connected ones. This explicitly encourages preferential attachment, yielding a training process where sparsely connected neurons are likely to be weeded out. 

\subsection{Attention Sparsification}
It is not obvious what topological sparsity based on preferential attachment would entail for attention. 
We conceive of a novel way of inducing such sparsity by leveraging the rich literature on group sparsity \cite{bach_2012, groupsparse_2017}.

Considering the connectivity of an input embedding neuron to the output neurons of an attention layer, it is evident that the $i^{\text{th}}$ embedding dimension only interacts with the $i^{\text{th}}$ row of the $Q$, $K$ and $V$ matrices. These interactions can be visualized as connections to the output neurons. However, even one non-zero entry in the $i^{\text{th}}$ row of $Q$, $K$, $V$ leads to the $i^{\text{th}}$ input neuron being connected to all output neurons. Hence, to remove the effect of this neuron on the output neurons, one needs to zero out the $i^{\text{th}}$ row in all three matrices. In other words, a group sparsity penalty, where each group is a row of the concatenated $A=[Q,K,V]$ matrix, is desired. Such a penalty encourages sparse rows to become more sparse as it tries to eliminate those rows by making them (almost) zero, again showcasing preferential behavior.

Rather than adding extra masking variables to implement preferential behavior, we leverage group sparsity and apply an $l_p^q$ norm penalty on the rows of $[Q,K,V]$, where $p=1$ and $q=0.5$. The $l_1^{.5}$ penalty was seen to be more robust to other choices in \cite{groupsparse_2017} as it leads to a sharp reduction in the parameter values belonging to a group. As such we add the following regularization, corresponding to the attention matrix at layer $l$, to the training loss, where $A^{(l)}$ is the concatenated $[Q,K,V]$ matrix:
\begin{align}
\label{eq:attn}
    R_{attn}^{(l)}(A^{(l)}) = \sum_{i=1}^d \sqrt{\sum_{j=1}^{3d} |A_{ij}^{(l)}|}
\end{align}
Note that the above constraint is applied across heads in the attention layer as it considers the entire $Q$, $K$, $V$ matrices (hence the inner summation over $3d$ entries). Additionally, while the standard $l_2$ group penalty induces weights within a group to be similar, this $l_1^{.5}$ group penalty allows sparsity patterns to be learned within a group. For example, in Figures \ref{fig:topo_sparsity_intro} and \ref{fig:topo_sparsity_attn}, while entire rows are often removed, we also observe that certain rows only exhibit sparsity in $Q$ and $K$ while leaving corresponding rows of $V$ dense, which is still valuable as it indicates that attention may not be required for those neurons/dimensions. 

\subsection{Head Pruning}
Unlike the attention and MLP sparsifications mentioned above, head pruning is done after each epoch as seen in Algorithm \ref{algo:neuroprune} (\textsc{NeuroPrune}). The main idea here is to remove heads in a layer that are similar to other heads and are hence deemed redundant. We want to remove as many heads as possible in order to get maximum sparsification. \textsc{NeuroPrune} accomplishes this by determining which heads are similar to many other heads, and then maintaining such heads while removing others. Note that similarity is not transitive, 
and thus removal of heads is not trivial. Algorithm \ref{algo:headprune} details these steps.

\textsc{NeuroPrune} removes heads that are \emph{dominated} by other heads, i.e., the dominated head is similar to only a subset of heads that the dominating head is similar to. The problem of keeping a minimum number of heads based on similarity can be mapped to the \textit{dominating set problem} \cite{domset}, where each head is a vertex and each edge indicates being similar. We want to find the minimum number of vertices such that they, along with their adjacent vertices, account for all the vertices in the graph. This problem is NP-Hard and our approach (detailed in Algorithm \ref{algo:dominating}) is a quadratic-time approximation to solve this, where it biases towards keeping latter heads in a layer. Since our algorithm also biases towards keeping vertices (heads) with high degrees, our head pruning scheme also elicits preferential behavior.

An important note to make is that, unlike previous methods \cite{michel_19,Li2021_DSP}, \textsc{NeuroPrune} does not prune according to head importance, but rather \textit{head redundancy}, and hence even important heads can get pruned. The experiments indeed show that the average head importance is quite high across the heads we eliminate. This can lead to more aggressive pruning and faster train times as witnessed in our experiments.

\subsection{\textsc{NeuroPrune}}
Algorithm \ref{algo:neuroprune} puts together the above regularizations and head pruning. Our fine-tuning procedure is very similar to the vanilla Stochastic Gradient Descent (SGD) methods \cite{large_scale_ml} that are typically used for training LLMs. In each epoch, the \textit{SGD($\cdot$)} term refers to running a single epoch of any SGD algorithm over a batched dataset. The key additions made in \textsc{NeuroPrune} are the MLP and attention regularizations, which appear in the objective being passed to \textit{SGD($\cdot$)}. Head pruning is done after each epoch of stochastic gradient descent in the inner \textbf{for} loop in Algorithm \ref{algo:neuroprune}.

\begin{figure*}[t]
\centering
    \includegraphics[width=.64\columnwidth]{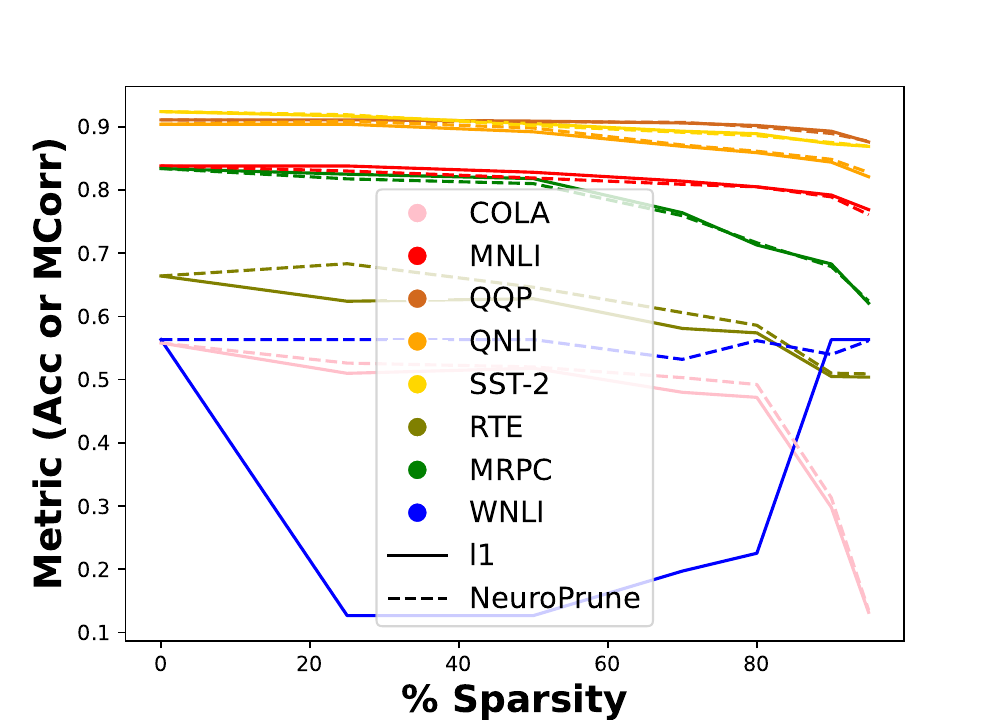} 
     \includegraphics[width=.65\columnwidth]{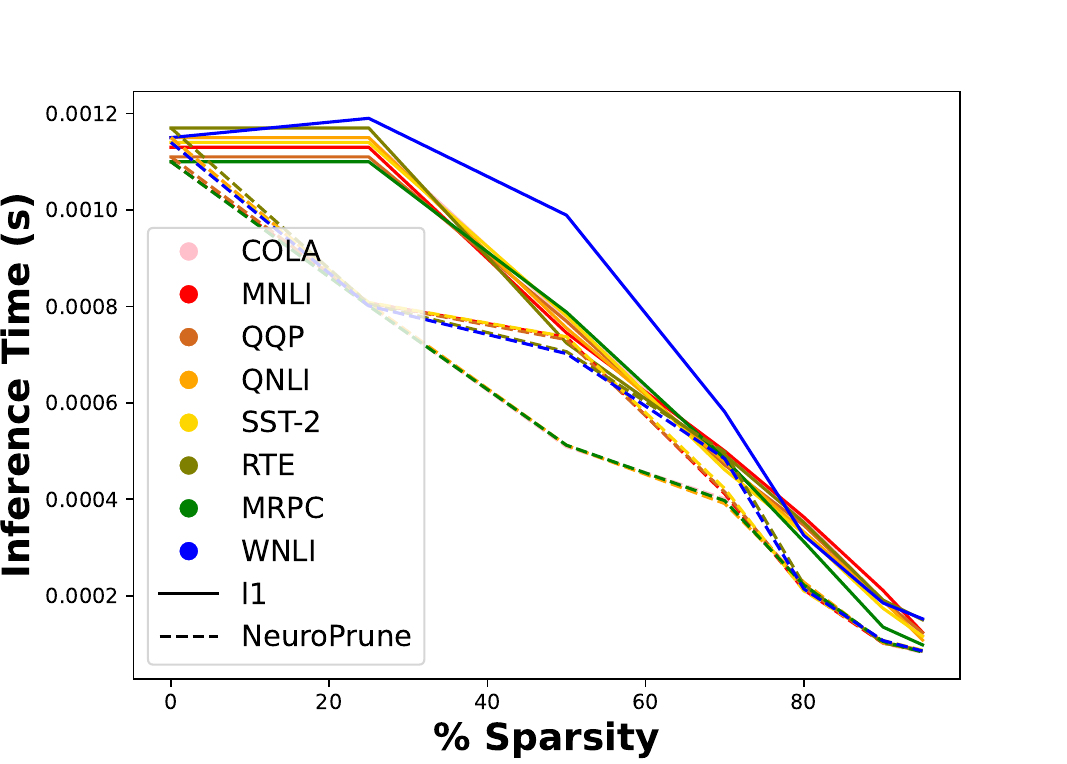}
      \includegraphics[width=.64\columnwidth]{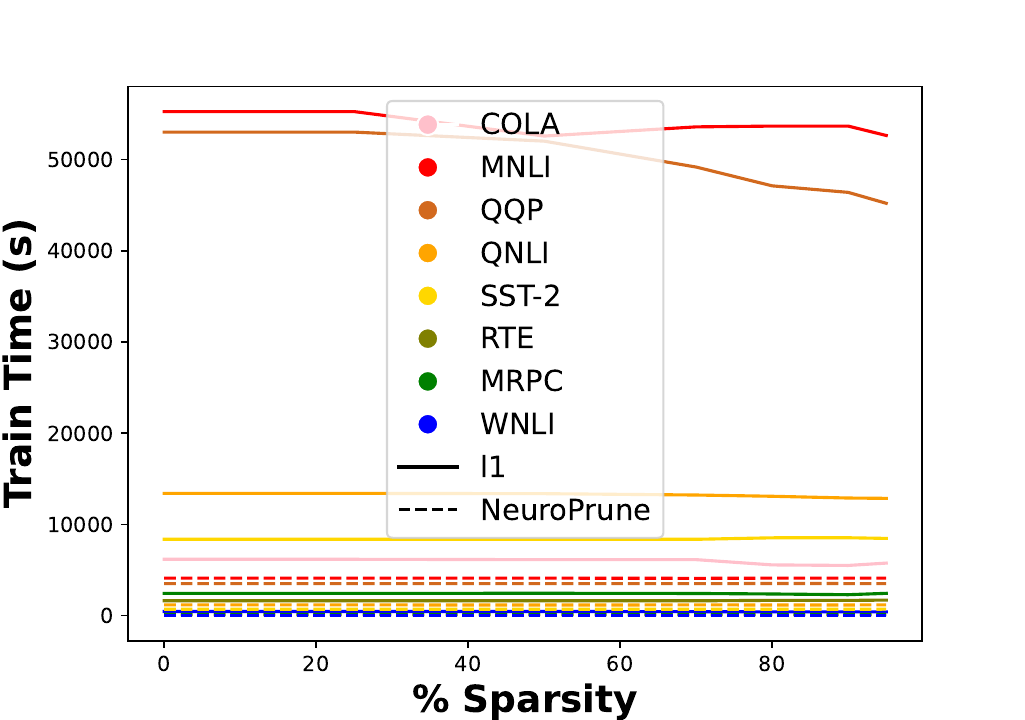}\\
      \includegraphics[width=.64\columnwidth]{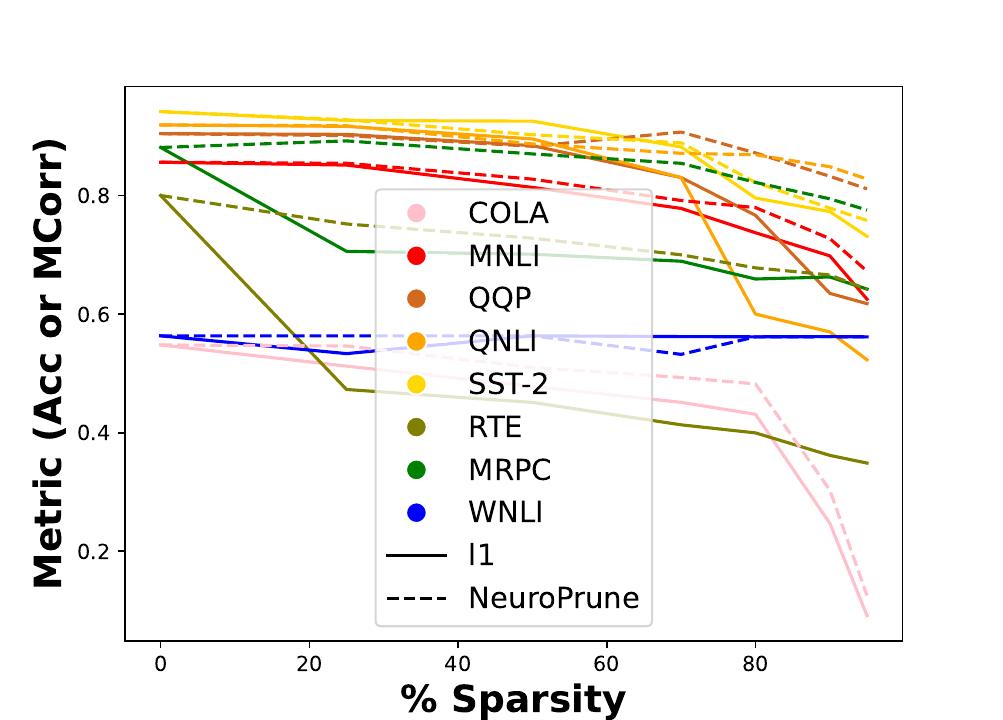}
      \includegraphics[width=.65\columnwidth]{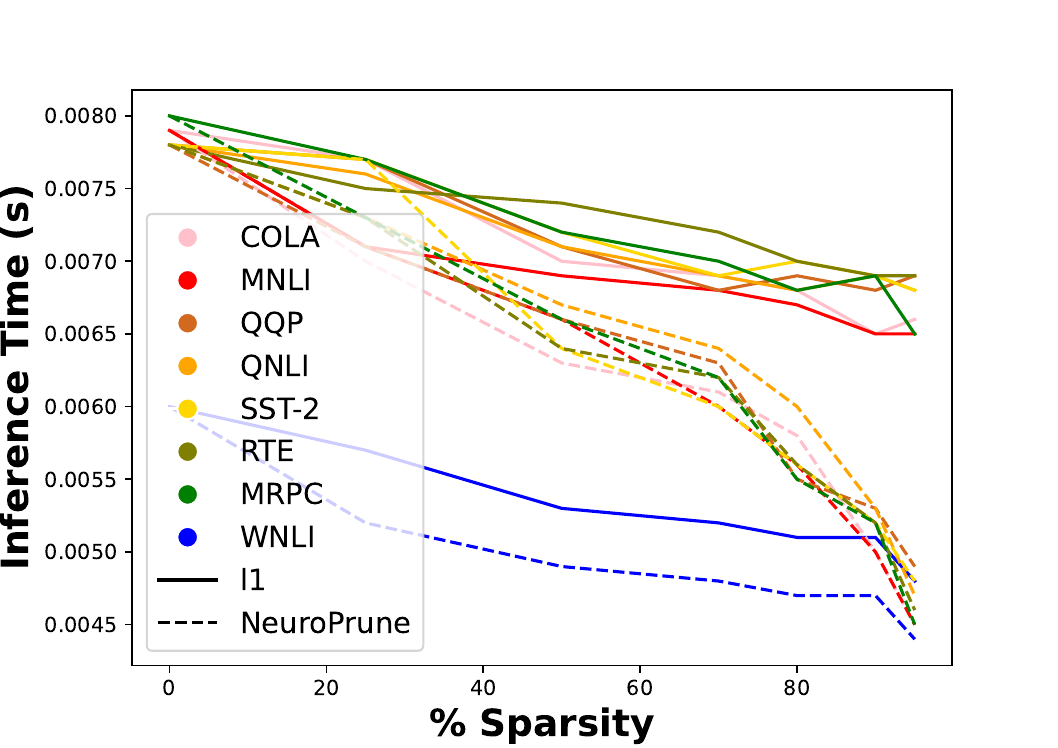}
      \includegraphics[width=.64\columnwidth]{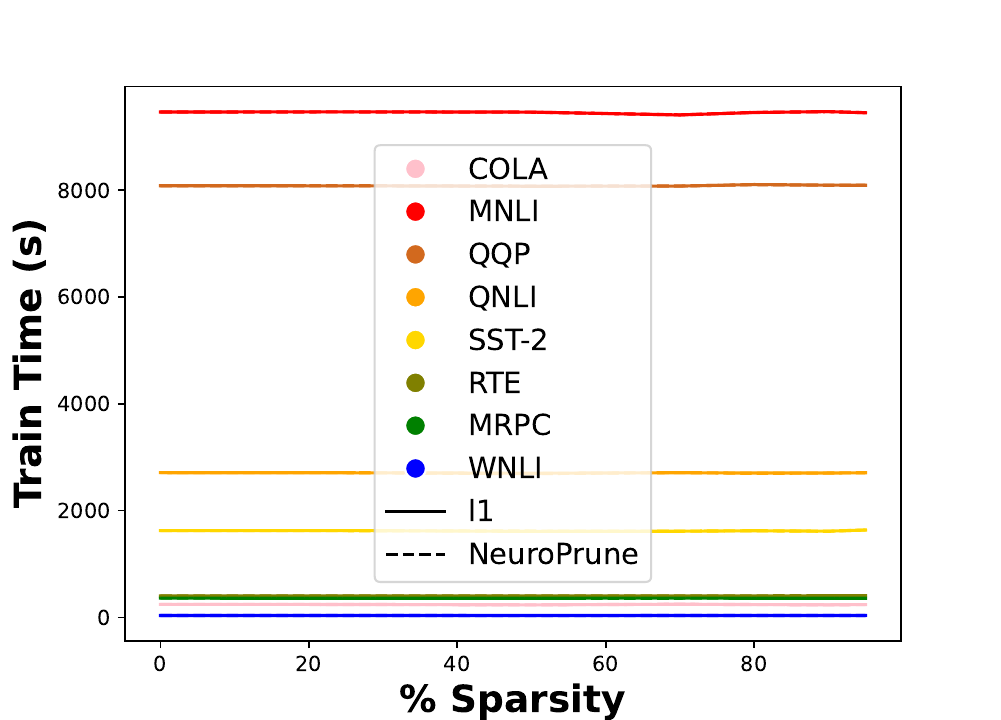}\\
      \includegraphics[width=.64\columnwidth]{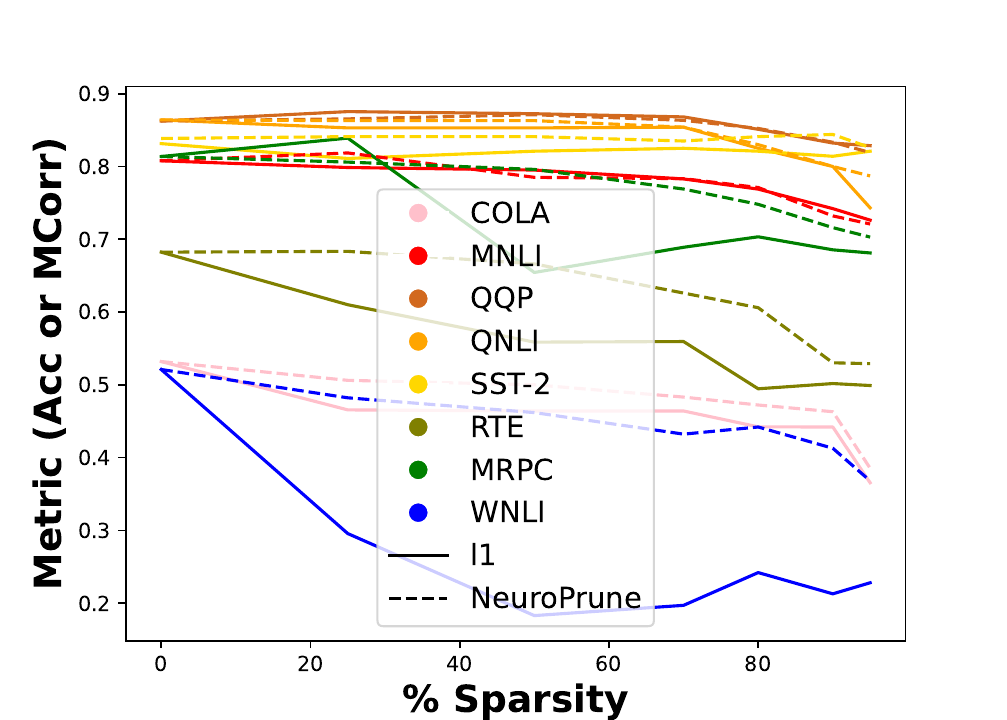}
      \includegraphics[width=.64\columnwidth]{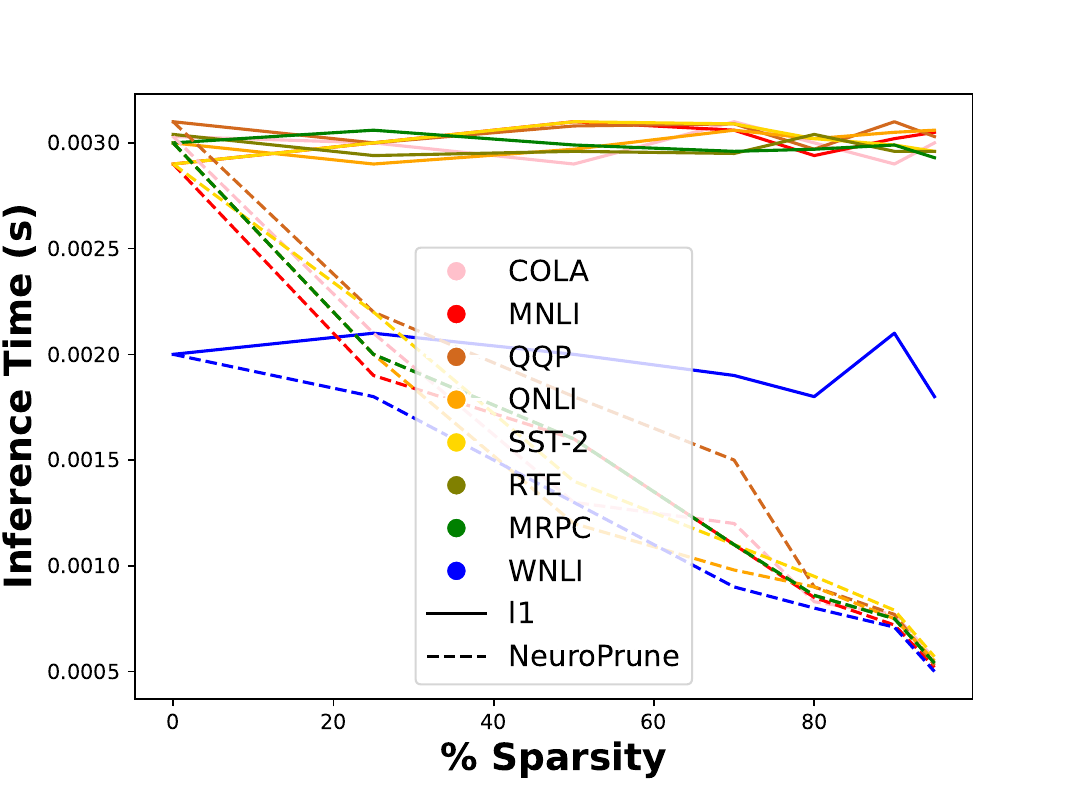}
      \includegraphics[width=.64\columnwidth]{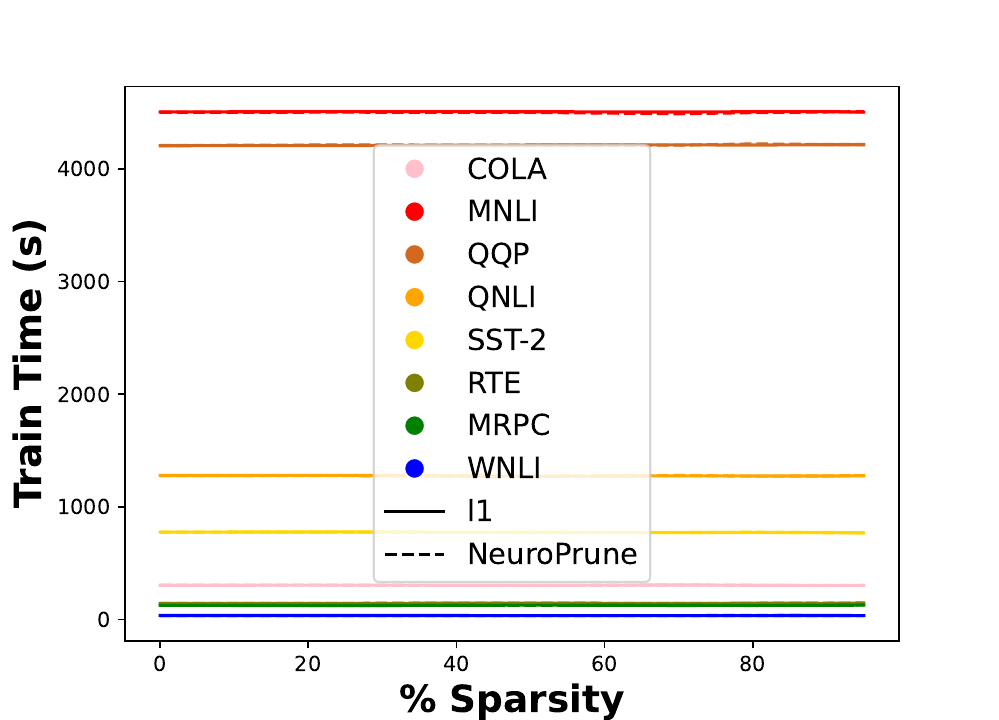}
    \caption{\small{Performance ($1^{\text{st}}$ column), inference time ($2^{\text{nd}}$ column) and train time ($3^{\text{rd}}$ column) for \textsc{NeuroPrune} and CoFI/$l_1$ on GLUE tasks at different sparsity percentages. The $1^{\text{st}}$, $2^{\text{nd}}$ and $3^{\text{rd}}$ rows correspond to BERT-base, T5-base and OPT-125m models respectively. In the $1^{\text{st}}$ row we see that \textsc{NeuroPrune} outperforms CoFI on the smaller GLUE datasets and is competitive on larger ones, with consistently better inference and train times. In the next two rows, we see that \textsc{NeuroPrune} is largely better than $l_1$ sparsity, especially at intermediate sparsities (25-80$\%$), with notable inference time gains and comparable train time. Qualitatively, similar results are obtained for T5-large and OPT-1.3b as seen in Figure \ref{fig:sparsity2} in the appendix.
    }}
    \label{fig:sparsity}
\end{figure*}

\setlength{\tabcolsep}{5pt}
\begin{table}[htbp]
 \centering
\caption{\small{NeuroPrune (NP) vs $l_1$ pruning on the CNNDaily summarization dataset using T5-base. FT stands for standard fine tuning. As an be seen we are most of the time better on rouge metrics and as well as inference time. The train times are similar. Best values for each sparsity \% ($s$) are bolded.
}}
\label{tab:cnn_sum}
\tiny
\begin{tabular}{cccccccc}
\toprule
\multirow{2}{*}{$s$}  & \multirow{2}{*}{Meth.}   &  $\uparrow$Rouge & $\uparrow$Rouge & $\uparrow$Rouge & $\uparrow$Rouge & $\downarrow$Inf. & $\downarrow$Train   \\
  &   &  $1$ & $2$ & L & Lsum & Time(s) & Time(s)   \\
 \midrule
$0$ & FT & $43.18$ & $20.47$ & $30.77$ &  $40.41$ &   $0.455$ &    $24603$ \\
\hline
\hline
\multirow{2}{*}{$25$} & NP & $\bf{43.07}$& $\bf{20.34}$ & $\bf{30.7}$ &  $\bf{40.31}$ &   $\bf{0.451}$ &    $\bf{24620}$ \\
\cline{2-8}
 & $l_1$ & $42.19$& $20.12$ & $29.29$ &  $39.33$ &   $0.454$ &    $24621$ \\
 \hline
 \multirow{2}{*}{$50$} & NP & $41.96$& $\bf{19.52}$ & $\bf{29.73}$ &  $\bf{39.2}$ &   $\bf{0.442}$ &    $24605$ \\
\cline{2-8}
 & $l_1$ & $\bf{42.18}$& $19.02$ & $29.29$ &  $38.65$ &   $0.451$ &    $\bf{24601}$ \\
 \hline
 \multirow{2}{*}{$70$} & NP & $\bf{41.6}$& $\bf{18.45}$ & $\bf{28.56}$ &  $\bf{37.93}$ &   $\bf{0.431}$ &    $\bf{24623}$ \\
\cline{2-8}
 & $l_1$ & $40.1$& $18.02$ & $27.51$ &  $36.63$ &   $0.441$ &    $24628$ \\
 \hline
 \multirow{2}{*}{$80$} & NP & $\bf{36.92}$& $\bf{16.78}$ & $\bf{26.29}$ &  $\bf{34.35}$ &   $\bf{0.427}$ &    $24614$ \\
\cline{2-8}
 & $l_1$ & $34.27$& $14.95$ & $25.11$ &  $32.79$ &   $0.437$ &    $\bf{24610}$ \\
 \hline
 \multirow{2}{*}{$90$} & NP & $\bf{33.92}$& $\bf{13.78}$ & $\bf{24.29}$ &  $\bf{31.35}$ &   $\bf{0.415}$ &    $\bf{24602}$ \\
\cline{2-8}
 & $l_1$ & $31.88$& $11.94$ & $23.18$ &  $29.22$ &   $0.422$ &    $24608$ \\
 \hline
 \multirow{2}{*}{$95$} & NP & $\bf{32.17}$& $\bf{13.72}$ & $\bf{23.66}$ &  $\bf{30.97}$ &   $\bf{0.406}$ &    $\bf{24610}$ \\
\cline{2-8}
 & $l_1$ & $30.25$& $11.21$ & $21.42$ &  $28.16$ &   $0.417$ &    $24611$ \\
\bottomrule
\end{tabular}
 
\end{table}

\begin{figure*}[t]
    \centering
    \includegraphics[width=.64\columnwidth]{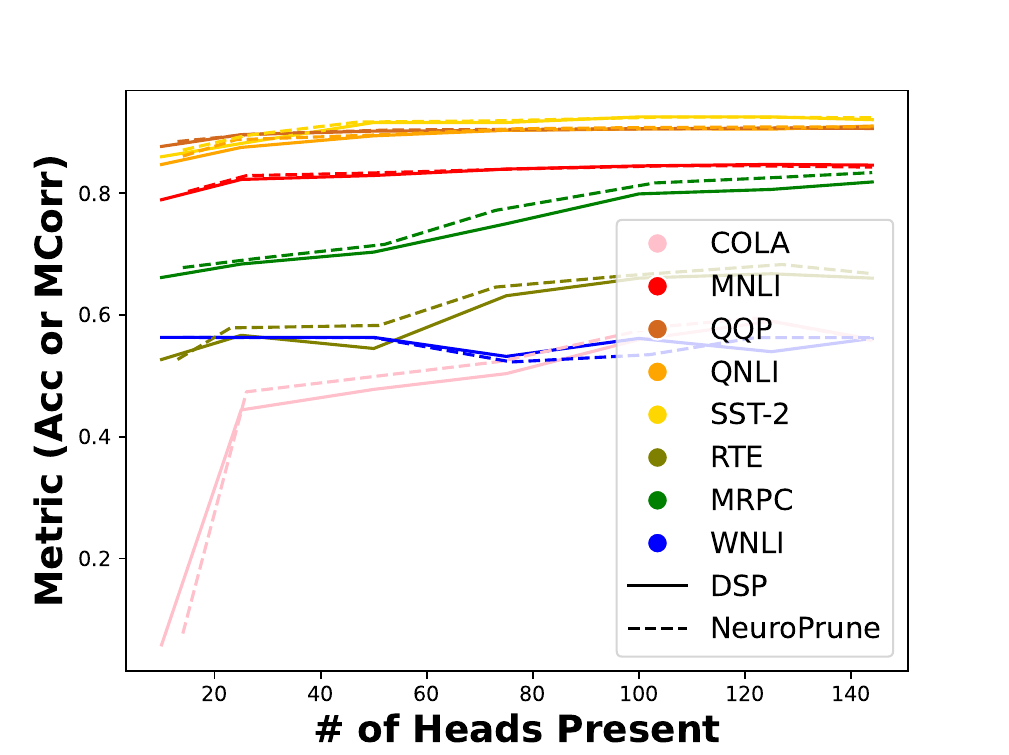}
     \includegraphics[width=.72\columnwidth]{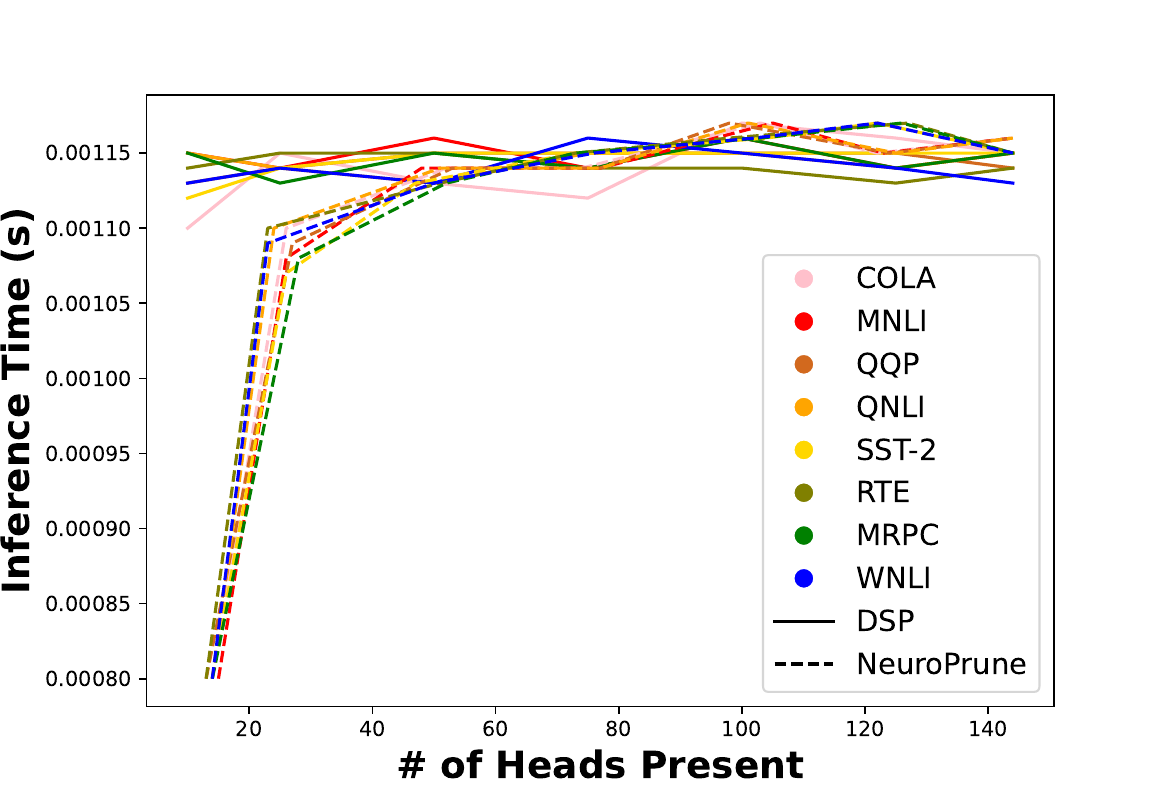}
      \includegraphics[width=.64\columnwidth]{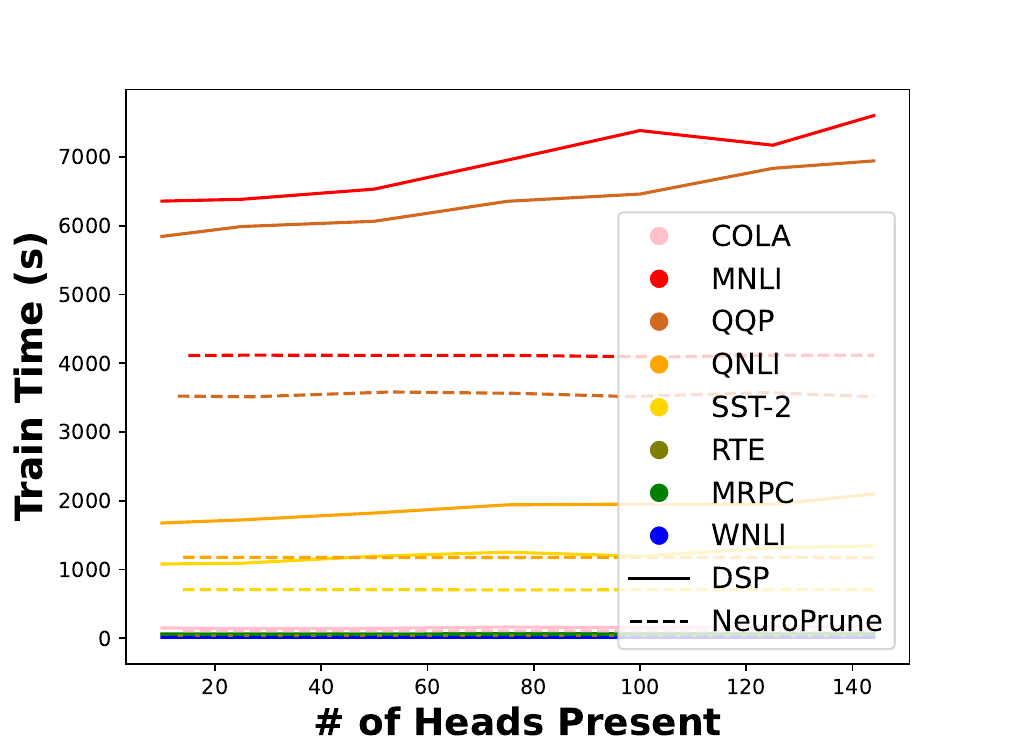}
      \caption{\small{Performance (left), inference time (center) and train time (right) for \textsc{NeuroPrune} and DSP on GLUE tasks, where different number of heads are present in a BERT-base model are shown above. \textsc{NeuroPrune} is better or similar (rarely worse) in performance to DSP on most datasets, where it is notably more efficient to train. Inference time is (slightly) improved when many heads are removed, however, the DSP code (simply) masks heads rather than explicitly pruning them like ours does and hence if these masked heads are removed the inference time of DSP might also improve as shown in their paper.}
    }
    \label{fig:headrem}
\end{figure*}

\section{Experiments}
\label{sec:exp}
We now test our method in two different settings: i) varying sparsity and ii) varying number of heads. In each setting, we run our method on the GLUE \cite{glue} tasks, where the dev set is used for testing\footnote{For MNLI we report the matched dev set accuracies.}. For i) we also test our method on the CNN/Daily Mail \cite{cnn} summarization task. For ii) we also run our method on a machine translation task for German to English on the IWSLT \cite{cettolo-etal-2014-report} dataset.

\noindent\textbf{Baselines and Models:} When varying sparsity we compare against CoFI \cite{Xia2022_CoFI}, which is a state-of-the-art (SOTA) method for inducing structured sparsity in LLMs. Since the author-shared code is for BERT (encoder) we compare with CoFI on BERT(-base) for GLUE tasks. We additionally implemented \textsc{NeuroPrune} for T5 (encoder-decoder) \cite{T5} and OPT (decoder) \cite{OPT} models. Since a CoFI implementation was not available for T5 and OPT, and would require modifying the architecture, we apply $l_1$ sparsity to the attention and MLP blocks of the transformer as a baseline. For summarization, we use a T5(-base) model and again $l_1$ as a baseline.

When varying the number of heads, CoFI does not provide an easy way to control for this number, and hence we compare against a specialized head pruning method called Differential Subset Pruning (DSP) \cite{Li2021_DSP}, another SOTA head pruning method. Here too, code is available for BERT, but not for T5 or OPT, and hence we compare NeuroPrune on BERT(-base) for the GLUE tasks. We do not show head removal results for the other models as there are no natural baselines like we had for sparsity ($l_1$). For machine translation, we use an 18-layer encoder-decoder model with 6 heads per layer as done in \cite{Li2021_DSP}. 

\noindent\textbf{Metrics:} For performance we report accuracy (Acc) for the GLUE datasets except COLA where Mathew's Correlation (MCorr) is a standard metric, Rouge for summarization and BLEU scores for machine translation. We also report (average) inference and train times. Please note that by train time we mean the total time taken to train the model using dynamic sparsity and inference time is the time taken by the model to process a test example based on the sparse model that has been learned.

\noindent\textbf{Hardware:} All experiments were conducted on an NVIDIA A100 GPU with 40 GB memory.

\noindent\textbf{Setup Details:} NeuroPrune results were obtained by varying $\alpha$ and $\beta$ from $10^{-7}$ to $0.1$ in multiples of $10$. The $l_1$ penalty parameter also took these values. $\theta$ took values in $\{0.15, 0.2, 0.25\}$ for the head removal experiments where the default was set to $0.15$ for the GLUE and summarization tasks. It was sufficient to run NeuroPrune for a single epoch for the larger GLUE datasets (viz. MNLI, QQP, QNLI and SST2) and the summarization task, while for the other smaller GLUE datasets we ran it for $3$ to $5$ epochs. $\epsilon$ for MLP sparsification was set to $10^{-4}$. 
CoFI finetunes the model before it starts pruning. For smaller GLUE datasets, finetuning before pruning epochs is $4$ and total epochs is $100$, while for larger datasets, these numbers are $2$ and $20$ respectively. All the other parameters were unchanged.  To make the comparison fair, we turned off the distillation option. DSP runs 3 epochs of joint (finetuning and mask learning) training. For machine translation, NeuroPrune results were obtained with $\alpha\in\{0.005, 0.01, 0.05, 0.1, 0.25, 0.5\}$, $\beta\in\{0.05, 0.1\}$ and $\theta$ varying from 0.1 to 0.4 in increments of 0.02. $\epsilon$ was set as in GLUE. Since there was no pretrained model, 15 epochs of pretraining were done followed by 15 epochs of NeuroPrune finetuning for a total of 30 epochs. DSP joint training was also done with a total of 30 epochs.

\subsection{Varying sparsity percentage}
We report results for sparsity percentages of $25$, $50$, $70$, $80$, $90$ and $95$ for GLUE and summarization.

\noindent\textbf{GLUE:} In Figure \ref{fig:sparsity} we see the behavior of our method on GLUE datasets for BERT, T5, and OPT models. We see that our method is always competitive with CoFI, outperforming it on the smaller datasets. This is possible because we do not add extra variables to the model, which when coupled with the topological constraints results in stabler performance. The structured sparsity also gives improved inference times and the train time is much lower, since not only do we need to run our method only for a few epochs, but the time per epoch is the same as standard fine-tuning. CoFI on the other hand, as recommended in \cite{Xia2022_CoFI}, requires many epochs of running to reach specified sparsity levels as it first fine tunes and then prunes.

For T5 and OPT we compare with $l_1$ sparsification. As can be seen, \textsc{NeuroPrune} is better than $l_1$ in most cases w.r.t. performance, especially for in-between sparsities. We believe this happens because \textsc{NeuroPrune} can choose between attention or MLP ($\alpha$, $\beta$ parameters), on which to sparsify more when optimizing performance even though the constraints are structured. We also see benefits in inference time, again possibly because of the structured sparsification. Train times are similar as we run $l_1$ for the same number of epochs as \textsc{NeuroPrune} and its per-epoch time is similar to that of standard fine-tuning. Performance with varying sparsity on T5-large and OPT-1.3b are seen in Figure \ref{fig:sparsity2} in the appendix.

\noindent\textbf{Summarization:} For summarization we see similar qualitative behavior of \textsc{NeuroPrune} vs $l_1$ sparsity for the T5 model. \textsc{NeuroPrune} is best on the Rouge metrics and its inference time is also slightly better. The train times are again similar.

\begin{figure}[htbp]
    \centering
    \includegraphics[width=0.49\columnwidth]{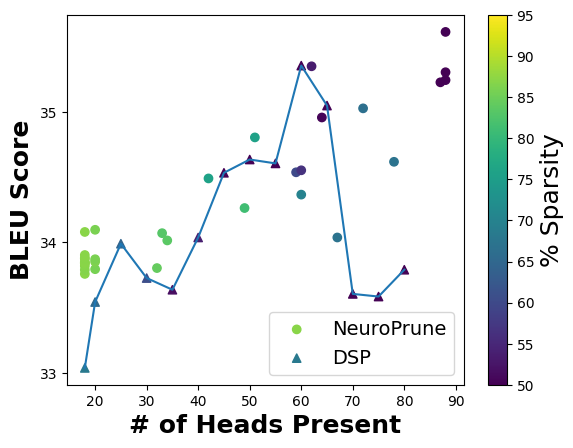}
     \includegraphics[width=0.49\columnwidth]{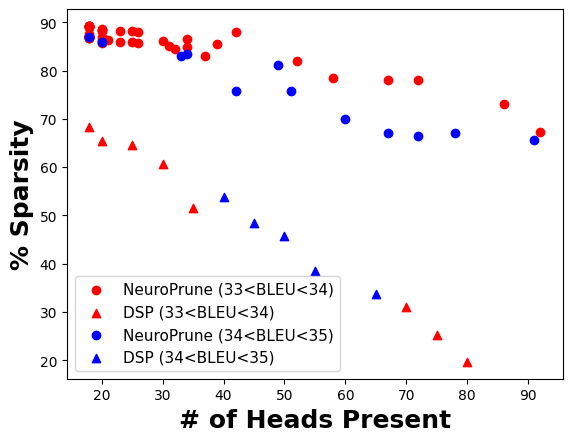}
      \caption{\small{Performance (left) and Efficient Frontier (right) for \textsc{NeuroPrune} and DSP on a German to English translation task. \textsc{NeuroPrune} consistently outperforms DSP across models with varying numbers of attention heads present and typically with much high levels of sparsity. The efficient frontier shows that for similar levels of performance and same number of heads, \textsc{NeuroPrune} finds much sparser models.}
    }
    \label{fig:mtl}
\end{figure}

\subsection{Varying number of heads}
Beyond sparsity, we now observe the behavior of \textsc{NeuroPrune} w.r.t. the number of heads pruned. We roughly keep $10$, $25$, $50$, $75$, $100$, $125$ and $144$ (i.e. all) heads. DSP is a strong competitor here.

\noindent\textbf{GLUE:} As can be seen we generally perform better or similar to DSP, but rarely worse. We believe this is because we have an additional knob of parameter sparsification, which can make heads similar as sparsity increases and given our novel redundancy-based head pruning algorithm we can effectively replace these heads with others keeping the overall performance of the model largely intact. The training time for our method is also significantly better since we can achieve the necessary head pruning in a few epochs, in addition to the fact that each epoch takes similar time as standard fine-tuning. The inference time is better, but that improvement may be reduced if the DSP code explicitly removed heads rather than just masking them\footnote{We use HuggingFace's \emph{prune\_heads()} function for this.}.

\begin{figure}[htbp]
    \centering    \includegraphics[width=.49\columnwidth]{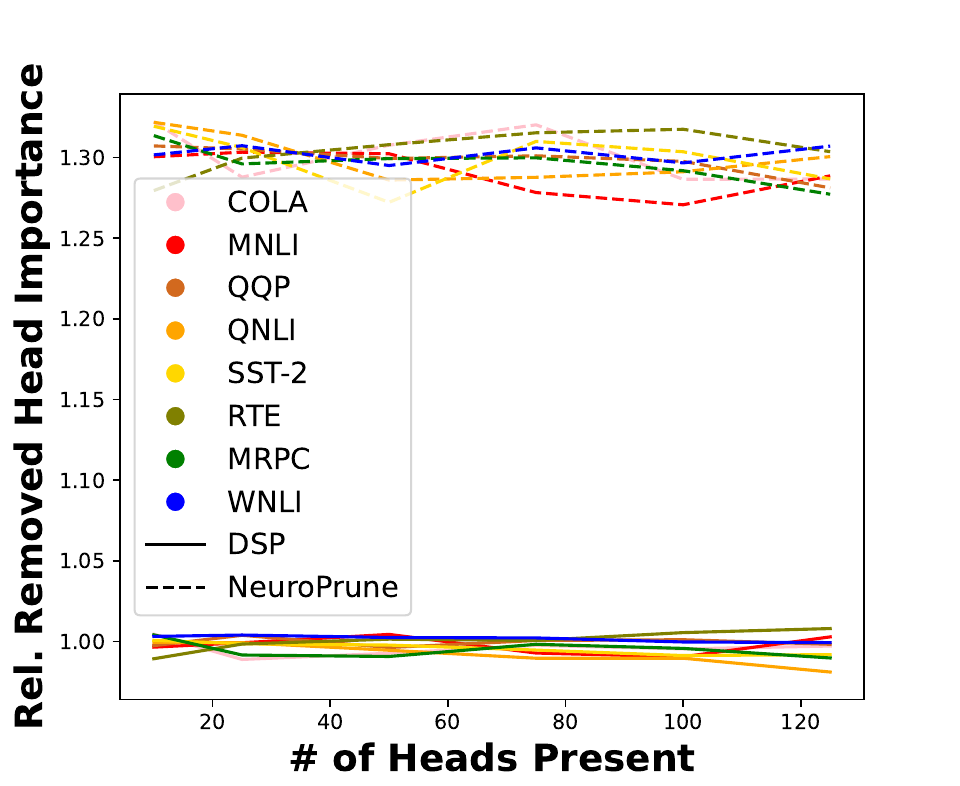}
\includegraphics[width=.49\columnwidth]{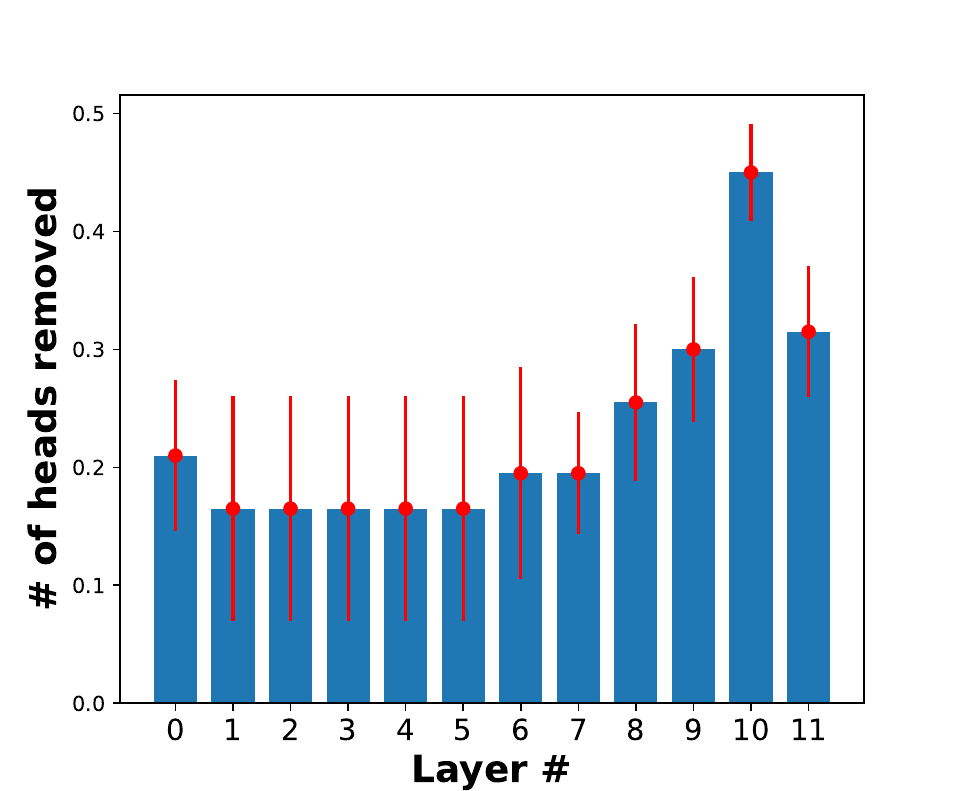}
    \caption{\small{Left, we see the relative importance of removed heads for NeuroPrune and DSP averaged across runs, where at least 10 heads are removed. We see that NeuroPrune removes more important heads than DSP does because of its redundancy based head elimination algorithm. Right, we see the (average) \# of heads removed per layer using NeuroPrune when fine tuning on the GLUE Benchmark datasets with a BERT(-base) model for cases where at least 10 heads are removed. As can be seen more heads are pruned from the later layers and the first one as compared to the middle layers.}
    }
    \label{fig:headimp}
\end{figure}

\noindent\textbf{Machine translation:} We see in Figure \ref{fig:mtl} (left) that \textsc{NeuroPrune} outperforms DSP on a machine translation task and with higher levels of sparsity. Figure \ref{fig:mtl} (right) illustrates efficient frontiers for both \textsc{NeuroPrune} and DSP; each curve offers a variety of models (varying sparsity and number of heads present) that achieve roughly similar performance. \textsc{NeuroPrune} clearly dominates DSP in terms of sparsity. DSP was run for varying numbers of heads present compared with varying sparsity parameters for \textsc{NeuroPrune} which generated more models. Unlike with GLUE tasks, we implemented \textsc{NeuroPrune} regularizations and head pruning within the DSP codebase which uses the fairseq toolkit \cite{fairseq} and used head masking for pruning. Since attention heads are still attached to the model, inference and train times between the methods on this task were similar; however note that the sparsity led to faster convergence, i.e., significantly better performance, for \textsc{NeuroPrune} when the number of heads present was greater than 70. \textsc{NeuroPrune} also has the advantage that it can be adapted to any architecture, whereas DSP must modify an architecture to include gates. Note that \citet{Li2021_DSP} show results from \citet{michel_19} (which significantly underperforms) and \citet{Viota2019_Head_Pruning} (which has similar performance but again requires direct architecture modification).

\subsection{Other Insights}

\noindent\textbf{Topological sparsity:} As seen in Figure 
\ref{fig:topo_sparsity_intro} (and Figures \ref{fig:topo_sparsity_attn}, \ref{fig:topo_sparsity_in}, \ref{fig:topo_sparsity_out} in the appendix) we see that our constraints try to eliminate neurons both in the MLP layers as well as the attention layers. This is observed as a row/column (sparsity) pattern in these matrices. Interestingly, sometimes only the $Q$ and $K$ entries in a row are sparsified, although the group sparsity constraint is applied to $QKV$ jointly.
The similarity in sparsification patterns for matrices $Q$ and $K$ directly reflects the symmetry of the operations they are jointly involved in, when computing the attention coefficients as inner products of \emph{each row} of $Q$ with \emph{each row} of $K$, compactly as $Q K^\top$,   
unlike $V$ which is simply a linear projection of the token embeddings. 
The MLP portions, as seen in Figure \ref{fig:topo_sparsity_intro}, exhibit preferential attachment with increasing sparsity which is consistent with brain functional networks.

\noindent\textbf{Importance of removed heads:} In Figure \ref{fig:headimp}(left), we see the \textit{relative importance of removed heads}, defined as the ratio between the importance of removed heads in a layer to the ones that remain averaged across all the layers. The individual head importances are computed as the sum of the absolute output dense weights that emerge from a head. We see that \textsc{NeuroPrune} can prune more important heads than DSP as it finds redundant heads, thereby sparsifying more aggressively, which is evidenced by the faster training time on GLUE tasks for similar levels of head removal.

\noindent\textbf{Head removal:} As seen in Figure \ref{fig:headrem} there is a high bias to keep the last head in each layer. This is because \textsc{NeuroPrune} prefers keeping later heads in a layer if it is similar to a set of heads that another head may be similar to, thus encouraging more modularity in a layer. In Figure \ref{fig:headimp}(right) in the appendix, we see that later layers, and often the first layer, lose more heads than the intermediate ones, which is consistent with \cite{Li2021_DSP}.
\section{Discussion}
This work shows that \textsc{NeuroPrune}, inspired by sparsity, modularity, and preferential attachment topology in brain functional networks, is competitive and sometimes even superior to other SOTA dynamic sparse training methods, even though it does not solely try to optimize performance. It is also more efficient than them in train time with speed-ups also seen in inference. Not to mention it is easily transferable across transformer architectures (i.e. encoder, encoder-decoder, decoder) as it does not require adding additional gating variables to the architecture or modifying the architecture in any way. Also note that since dynamic sparse training and post-pruning are complimentary to each other one can potentially apply both, that is, the latter after the former. Moreover, since dynamic sparse training approaches employ sparsity during training the benefits of sparsity in terms of storage and also sometimes computation are leveraged both during training and inference. In contrast, post-pruning methods apply pruning after (dense) network training, benefiting from sparsity at inference alone.

There are multiple avenues for future research. First, it would be interesting to combine our redundancy-based and head-importance pruning methods to produce even more aggressive and efficient pruning of heads. Second, the structured strategies we used for fine-tuning could also be tried during pre-training. Third, one could test the generalizability of models trained using \textsc{NeuroPrune} on related, but different tasks and measure if similar gains can be secured. We believe our work will spur further progress on efficient dynamic sparse training methods that are also performant.

\section*{Limitations}
All the datasets we considered were in English. Results may vary for other languages. We applied our method to three LLMs, but more architectures could be tested in the future not to mention more tasks. Our method although easy to adapt to different architectures while also being efficient, it does not allow the user to specify the exact number of heads one wants to prune and architectures are limited to Transformer based LLMs. This is implicitly a function of the threshold $\theta$ and the similarity of the pre-trained/fine tuned heads as well as the sparsity. We also have three hyper-parameters ($\alpha$, $\beta$ and $\theta$) that need to be specified for each run.

\section*{Ethics Statement}
Our work could be used to dynamically sparsify other LLMs and models, while fine tuning or pre-training. The sparsification may result in reduced alignment if the LLM is aligned to certain values and especially if those values are not encapsulated by the loss function that is used to fine tune using our method. So although one may use the method to create smaller models one has to be cognizant of what aspects may have been lost in the process. The method is easy to adapt to transformer based models and hence could possibly be widely used but improvements such as being able to specify number of heads etc. might be beneficial in future versions.


\bibliography{custom}
\bibliographystyle{acl_natbib}
\clearpage
\appendix
\section{Additional Figures}
\begin{figure}[htbp]
    \centering
    \includegraphics[width=\columnwidth]{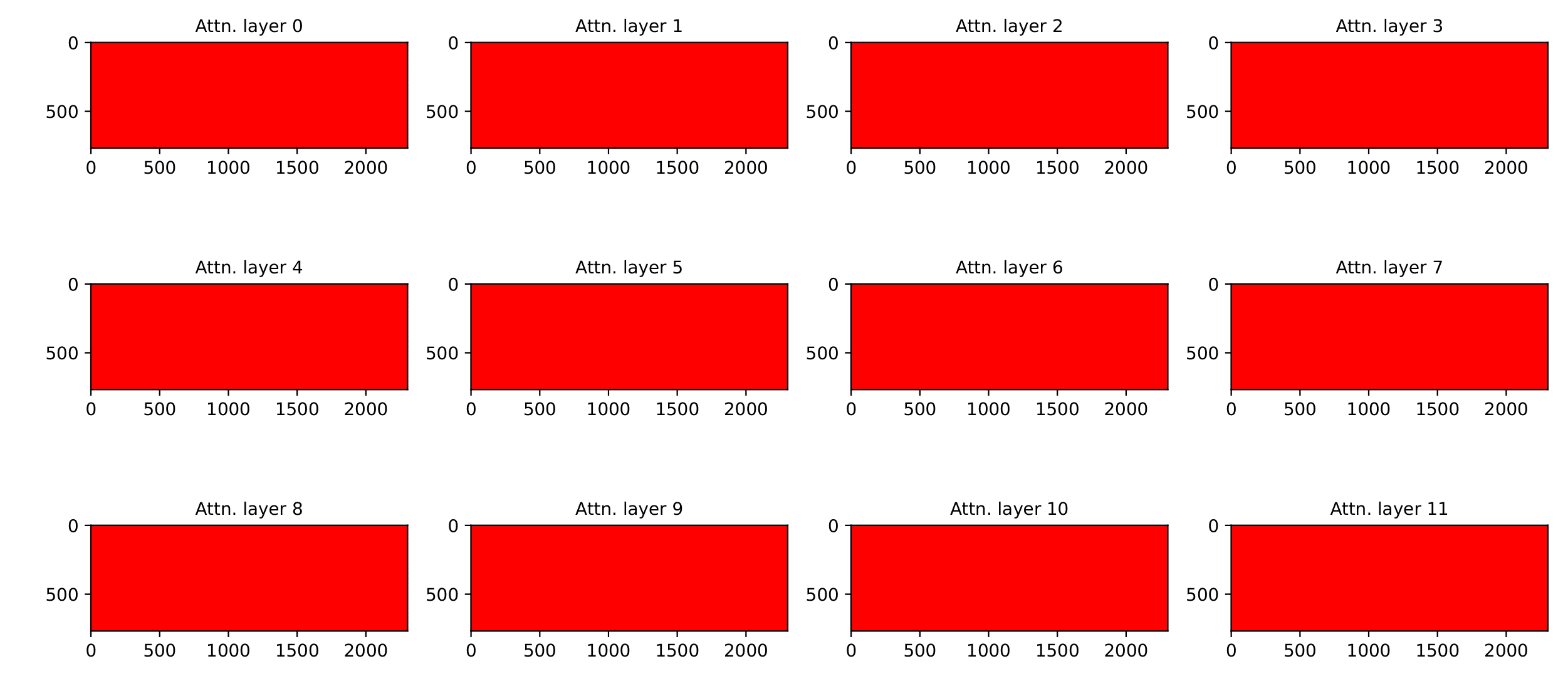}
    \includegraphics[width=\columnwidth]{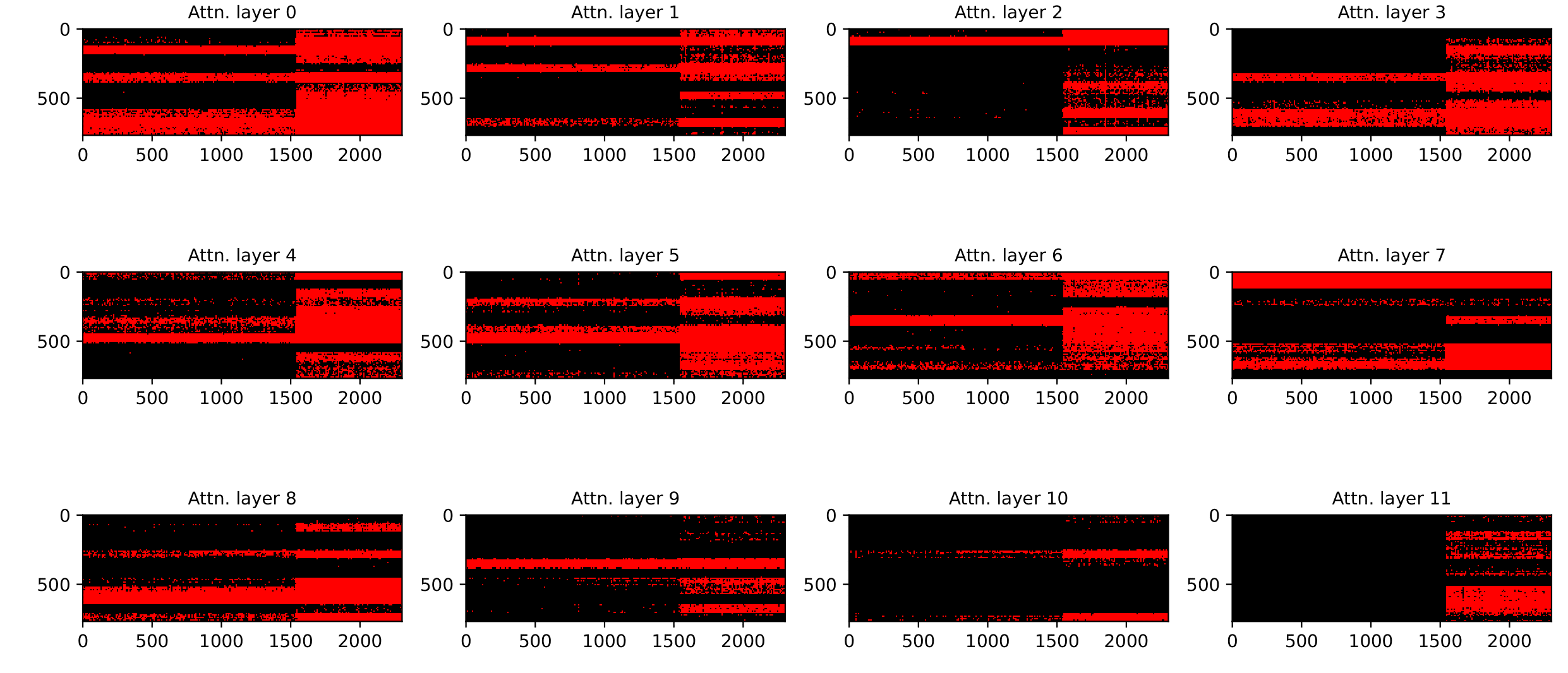}
    \includegraphics[width=\columnwidth]{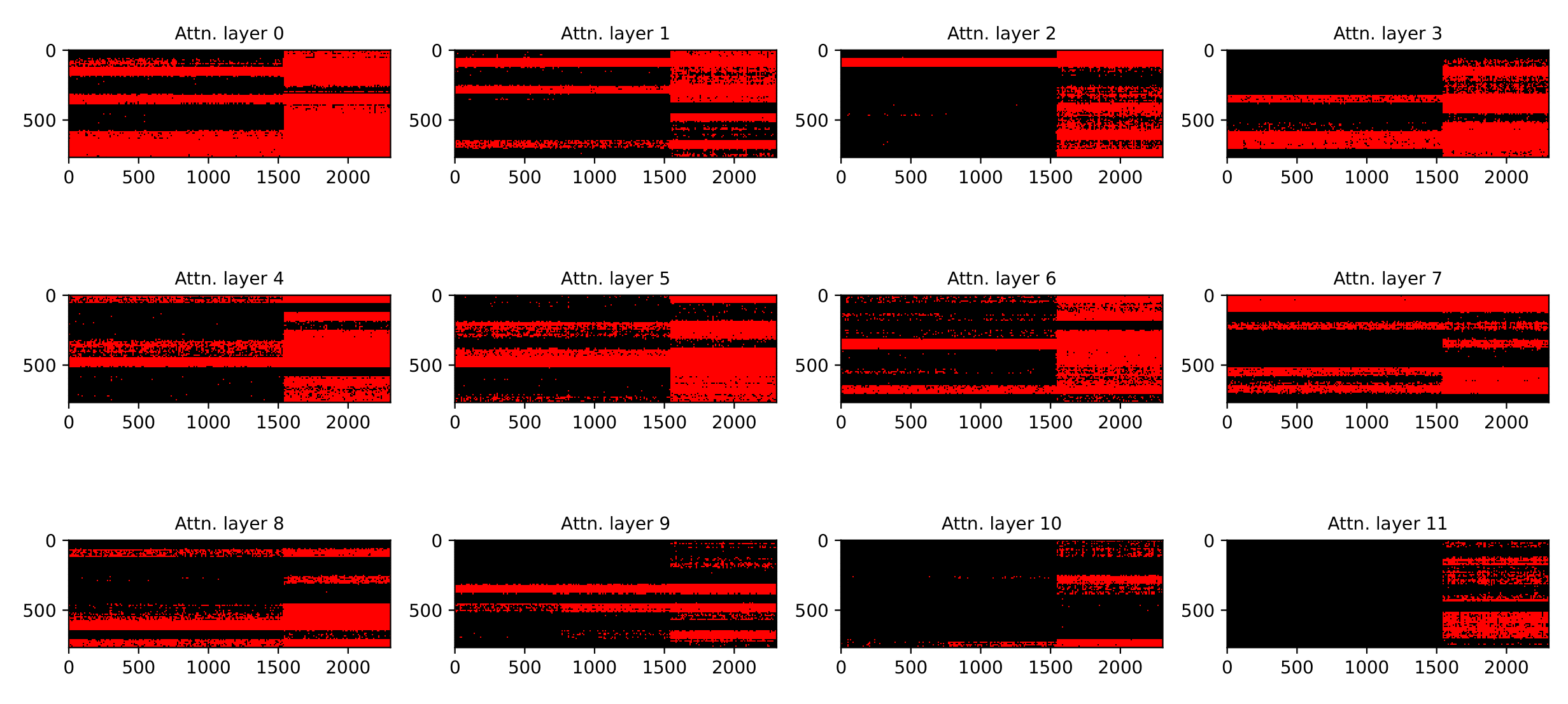}
    \includegraphics[width=\columnwidth]{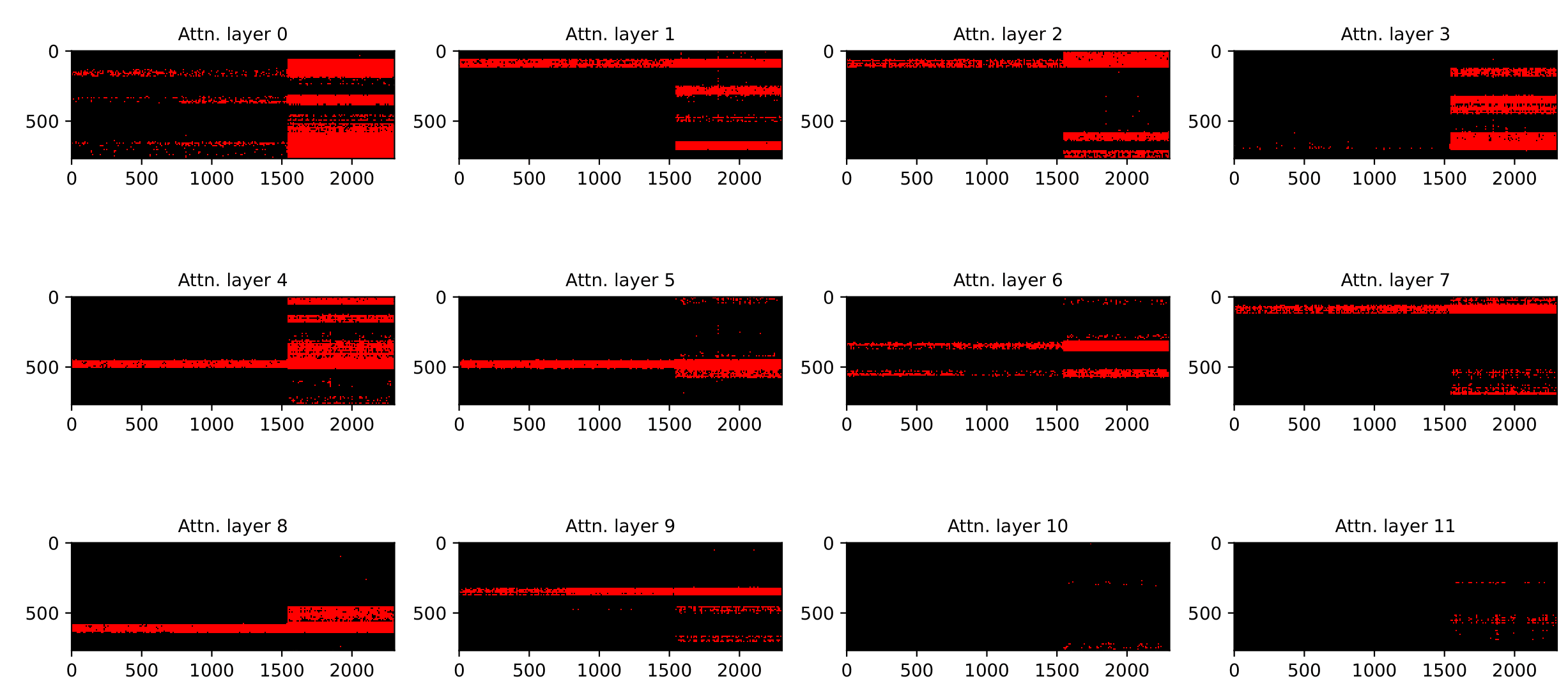}
    \caption{\small{Attention layers for BERT model where, top three rows correspond to standard fine tuning, the next sets of three correspond to $\approx 25\%$, $\approx 50\%$ and $\approx 90\%$ sparsity using NeuroPrune. As can be seen NeuroPrune encourages preferential attachment topology.}
    }
    \label{fig:topo_sparsity_attn}
\end{figure}

\begin{figure}[htbp]
    \centering
    \includegraphics[width=.6\columnwidth]{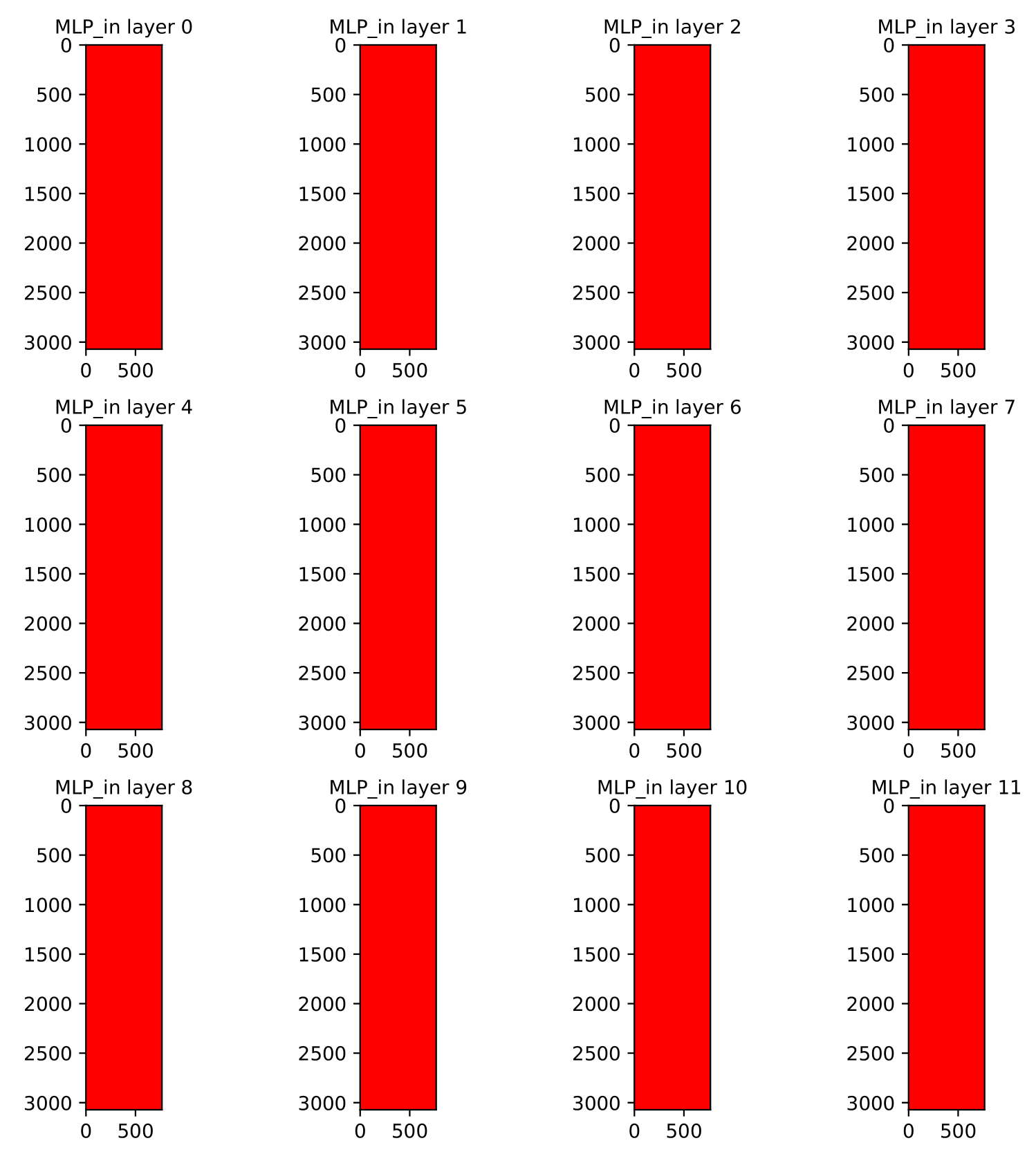}
    \includegraphics[width=.6\columnwidth]{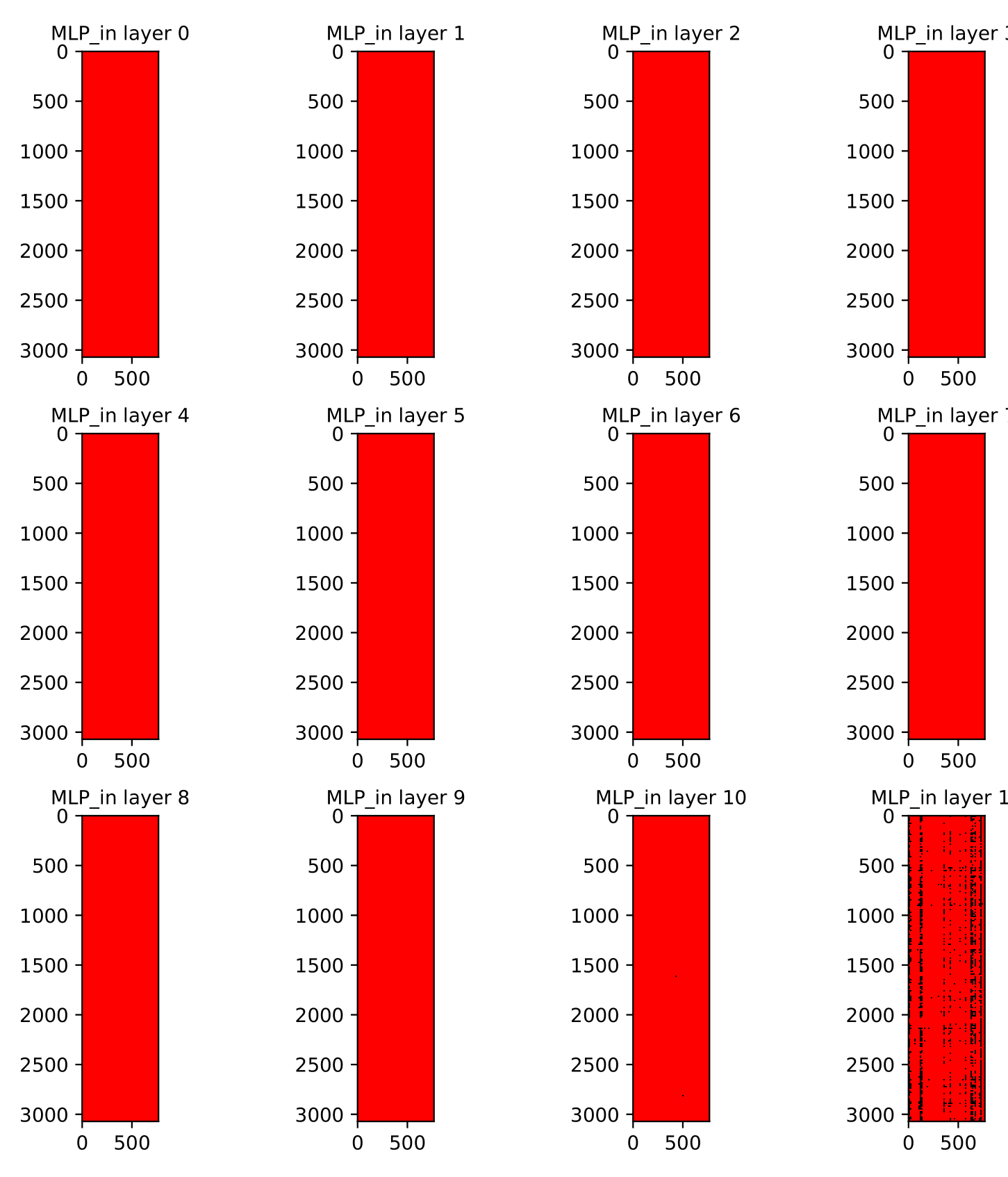}
    \includegraphics[width=.6\columnwidth]{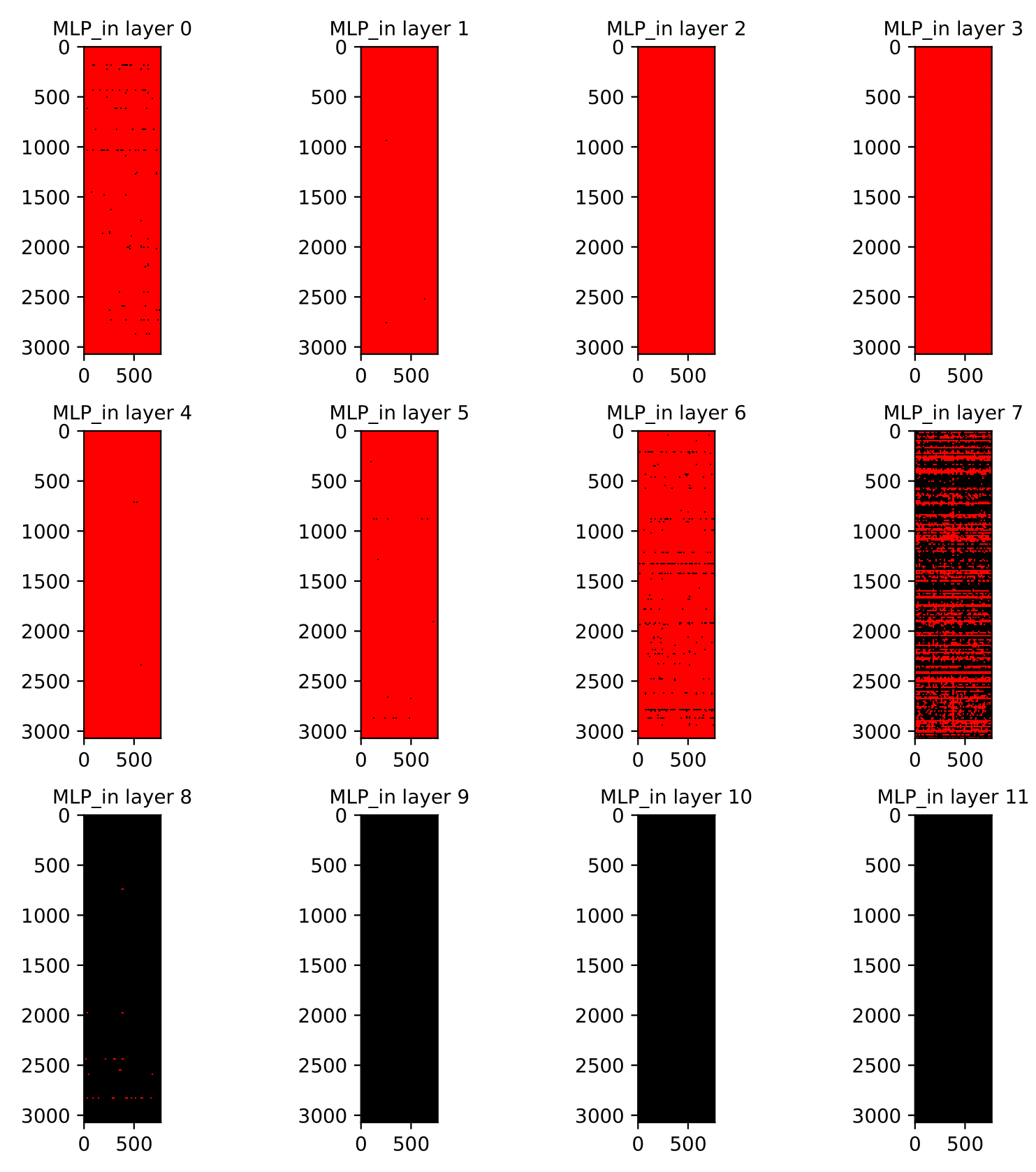}
    \includegraphics[width=.6\columnwidth]{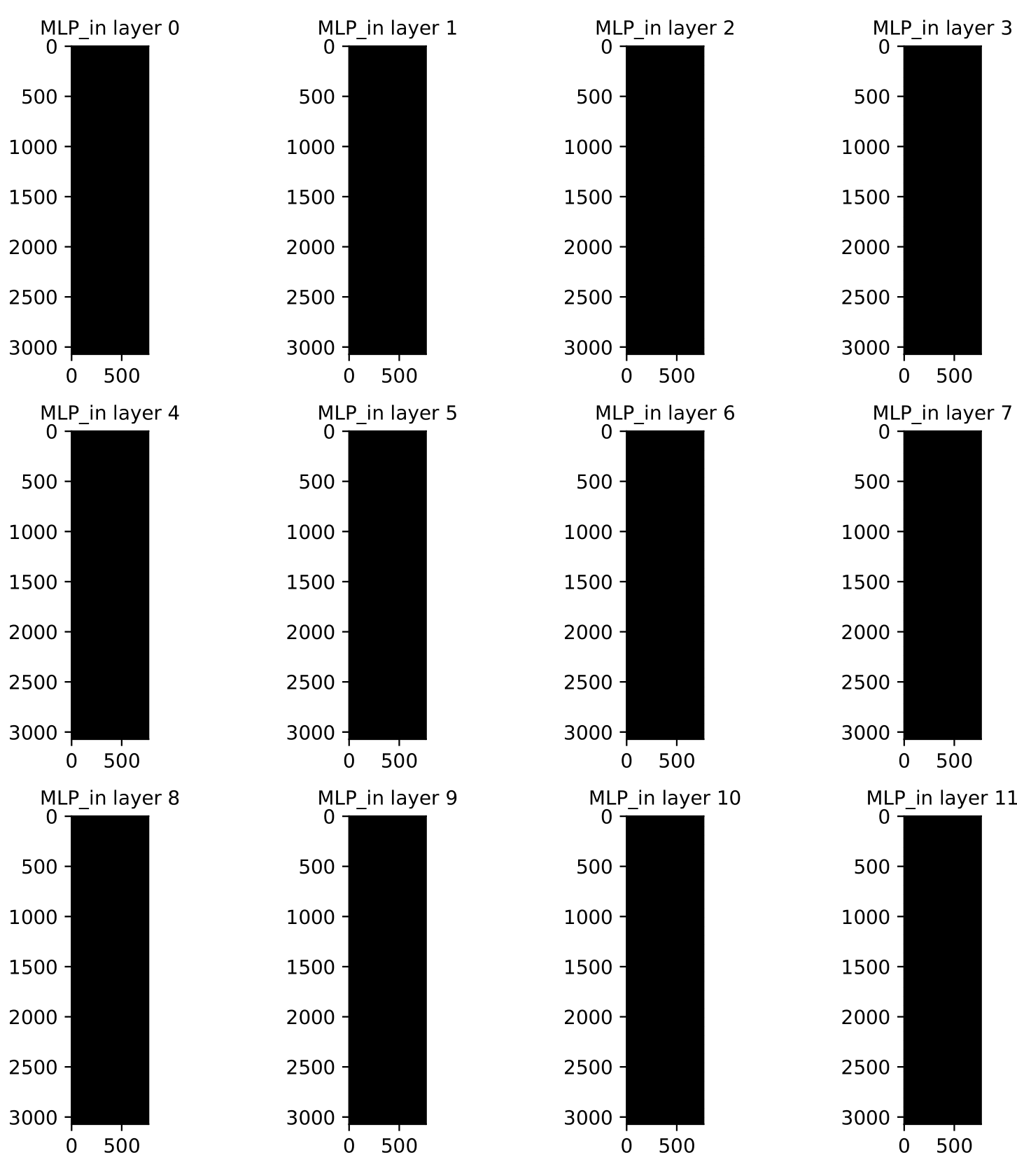}
    \caption{\small{MLP\_in layers for BERT model where, top three rows correspond to standard fine tuning, the next sets of three correspond to $\approx 25\%$, $\approx 50\%$ and $\approx 90\%$ sparsity using NeuroPrune. As can be seen NeuroPrune encourages preferential attachment topology.}
    }
    \label{fig:topo_sparsity_in}
\end{figure}

\begin{figure}[htbp]
    \centering
    \includegraphics[width=\columnwidth]{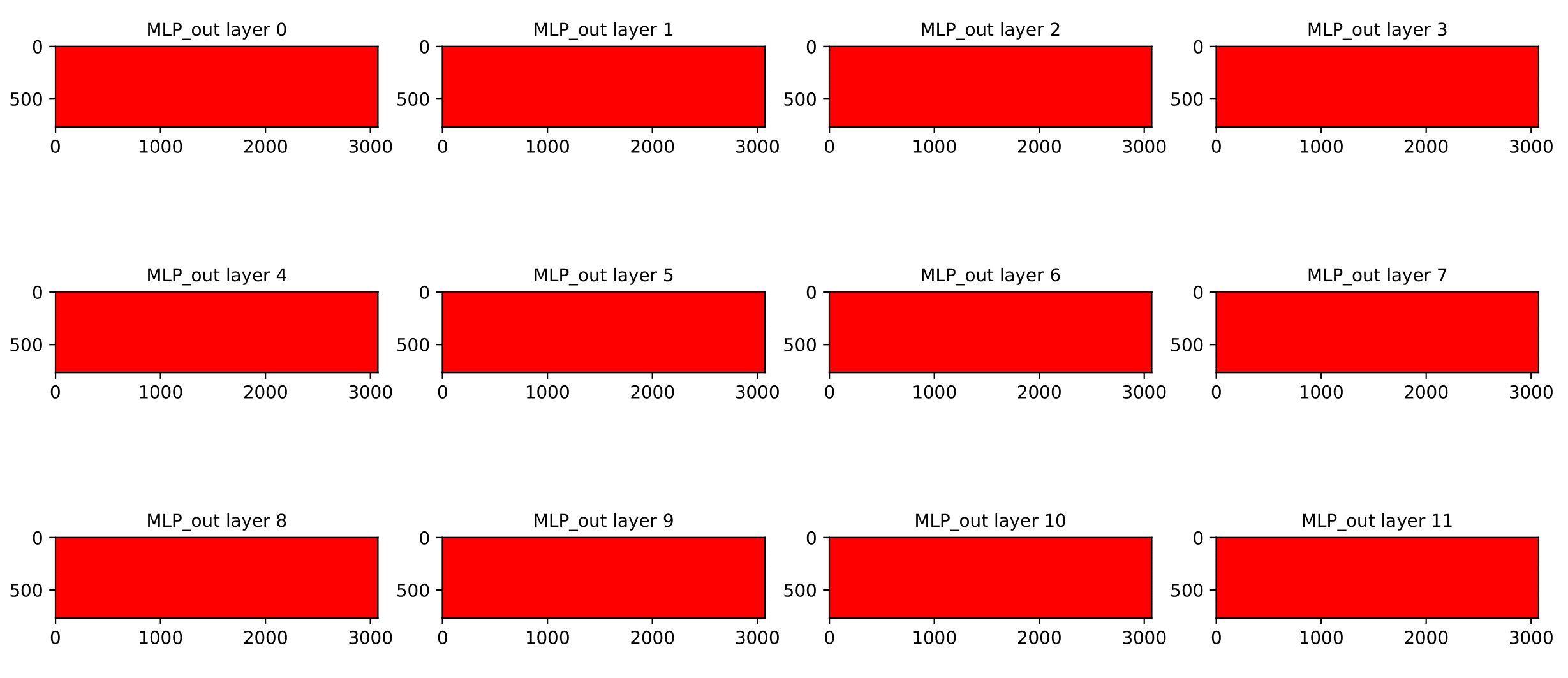}
    \includegraphics[width=\columnwidth]{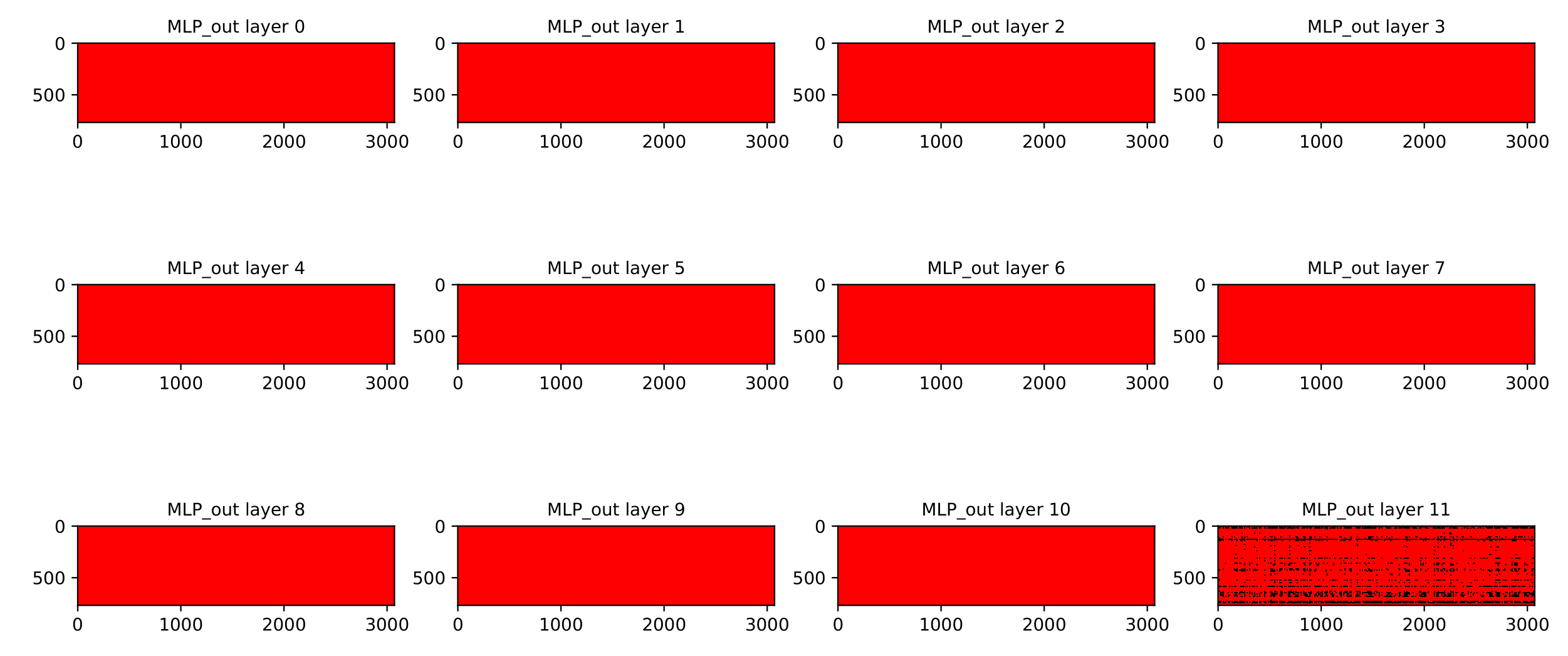}
    \includegraphics[width=\columnwidth]{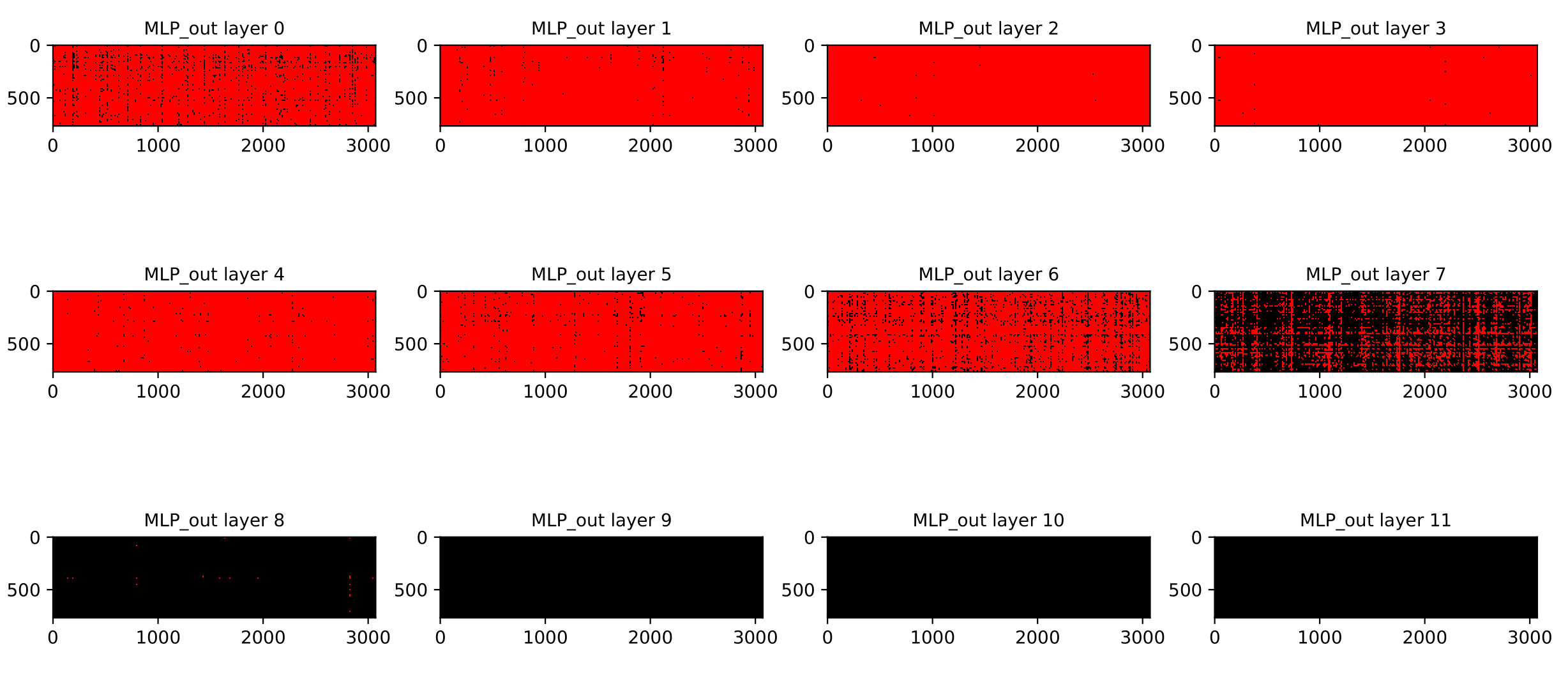}
    \includegraphics[width=\columnwidth]{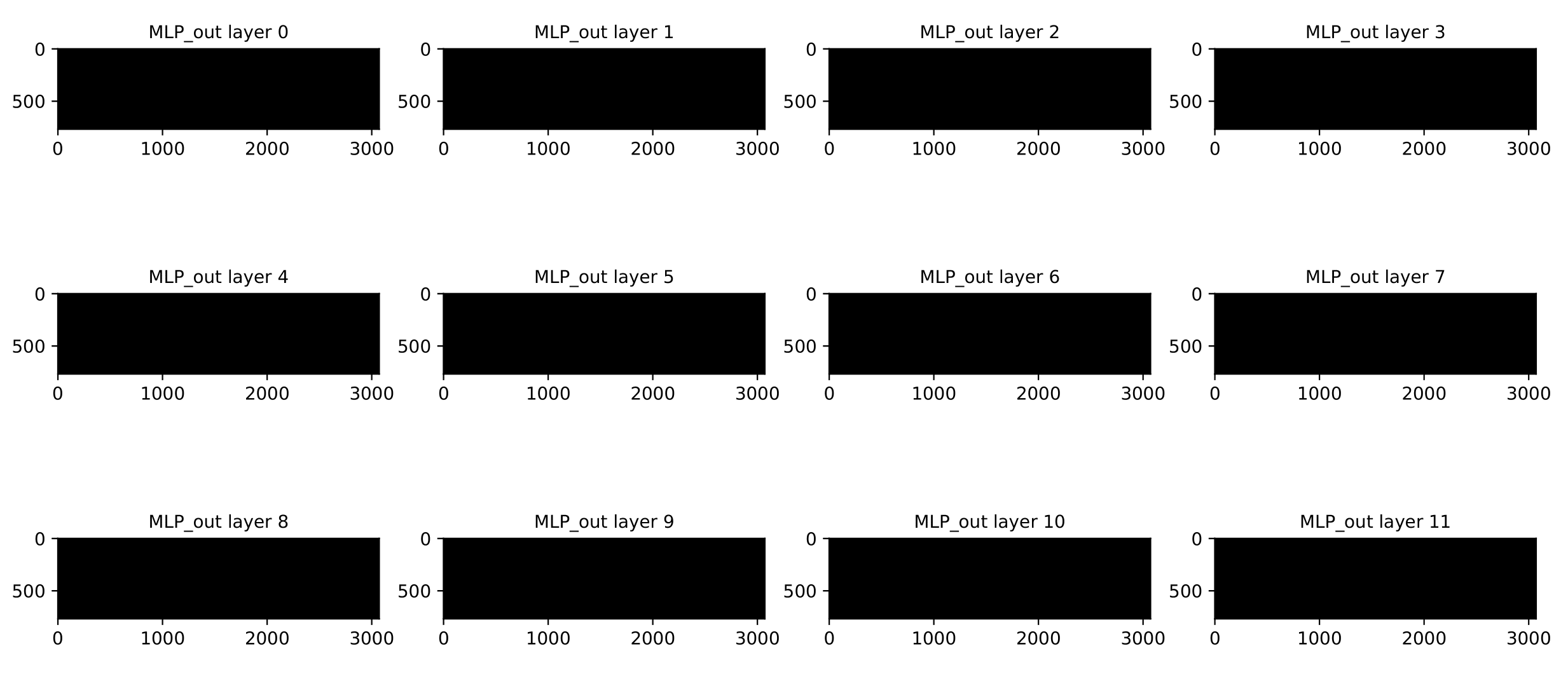}
    \caption{\small{MLP\_out layers for BERT model where, top three rows correspond to standard fine tuning, the next sets of three correspond to $\approx 25\%$, $\approx 50\%$ and $\approx 90\%$ sparsity using NeuroPrune. As can be seen NeuroPrune encourages preferential attachment topology.}
    }
    \label{fig:topo_sparsity_out}
\end{figure}

In Figures \ref{fig:topo_sparsity_attn}, \ref{fig:topo_sparsity_in} and \ref{fig:topo_sparsity_out} we sparsity patterns induced by NeuroPrune in a BERT-base model at different sparsity percentages on the SST2 dataset.

\begin{figure*}[t]
\centering
    \includegraphics[width=.64\columnwidth]{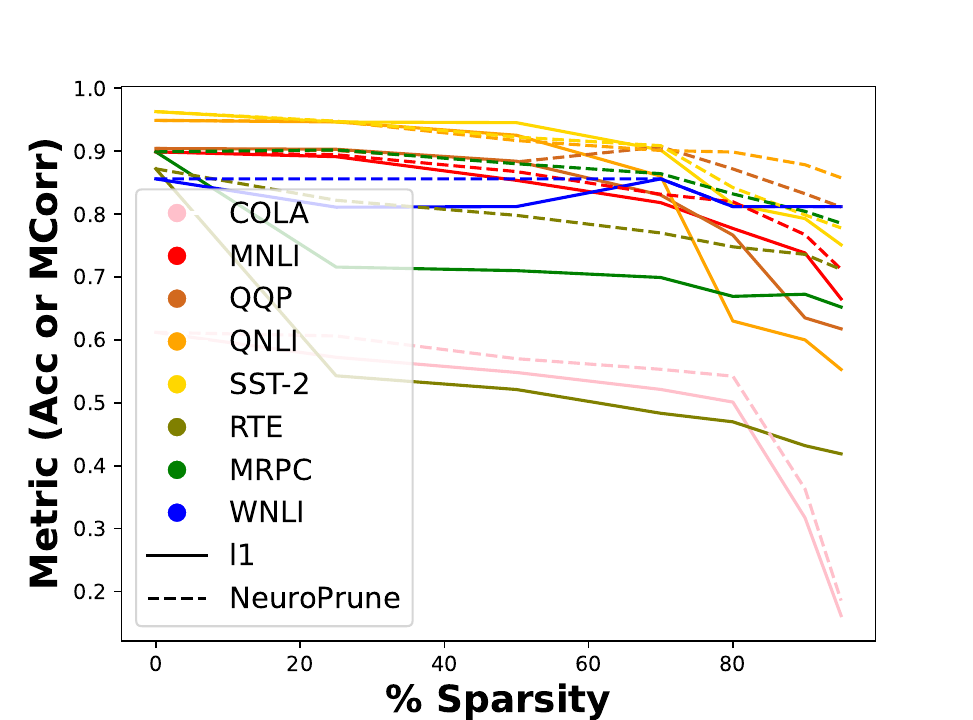}
      \includegraphics[width=.64\columnwidth]{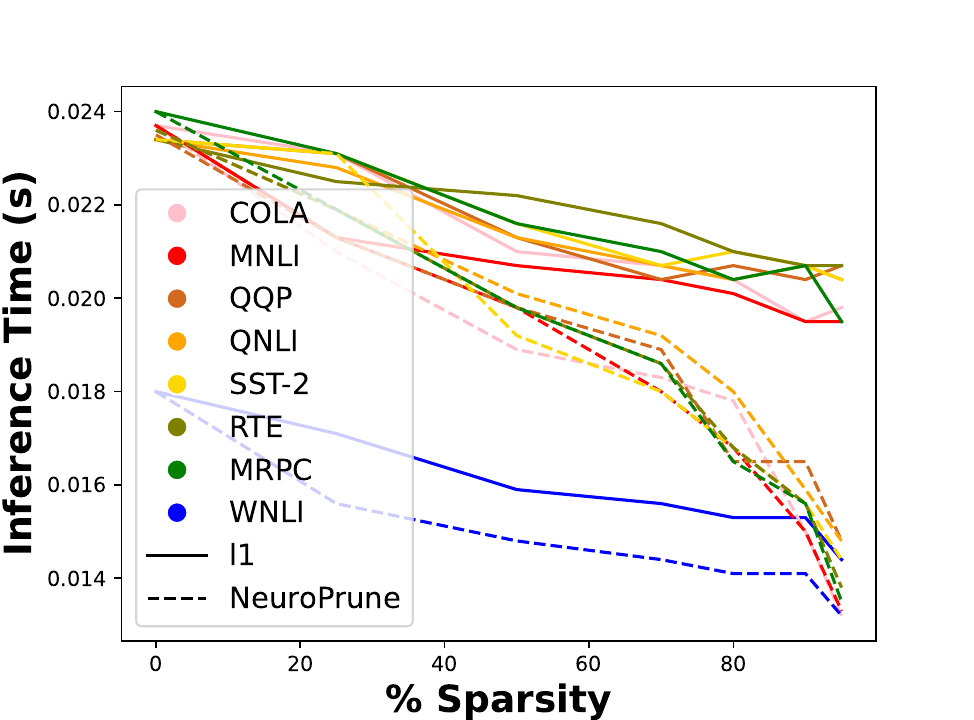}
      \includegraphics[width=.64\columnwidth]{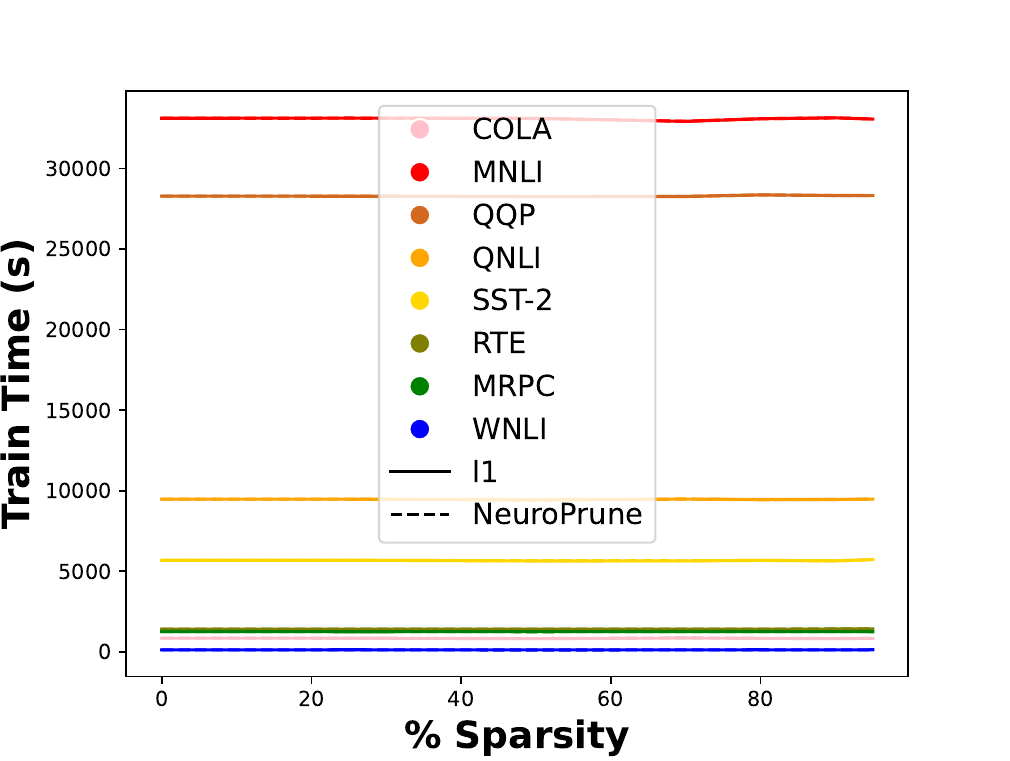}\\
      \includegraphics[width=.64\columnwidth]{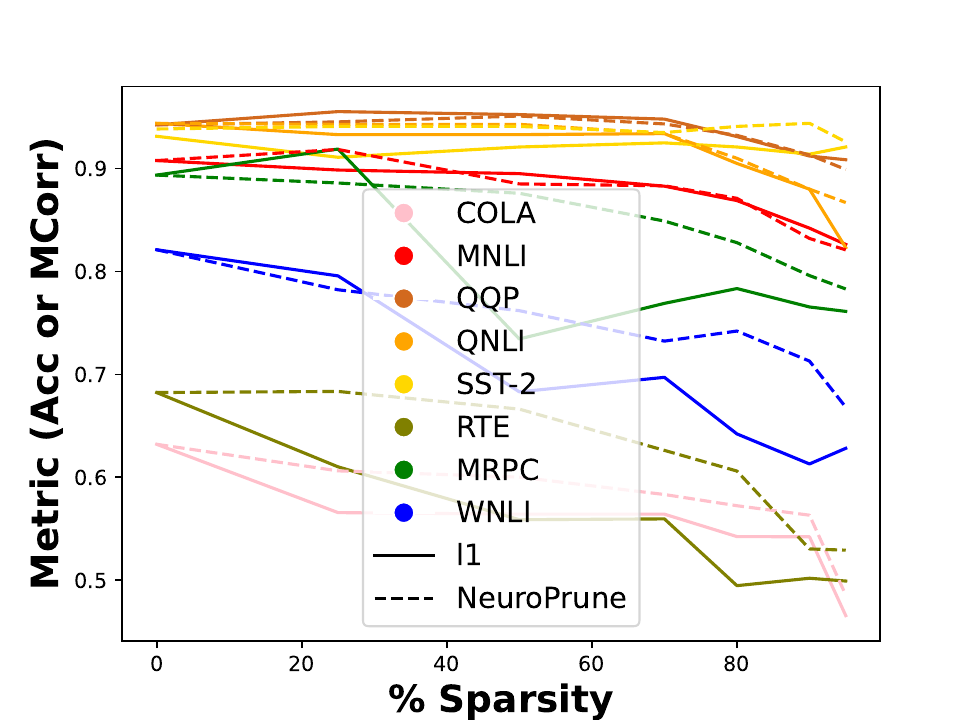}
      \includegraphics[width=.64\columnwidth]{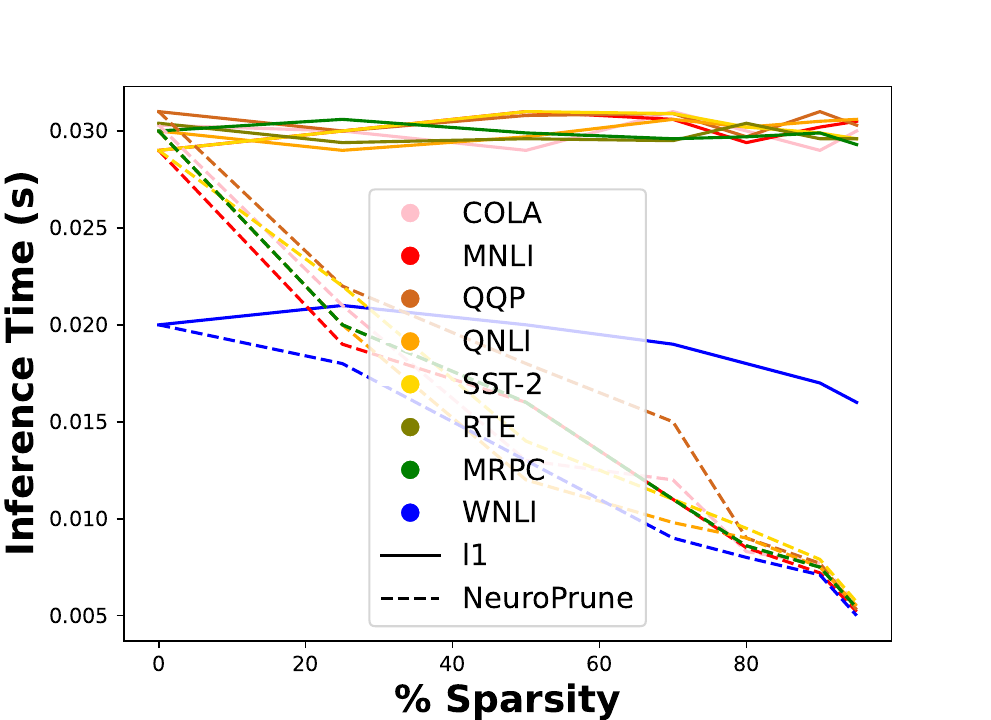}
      \includegraphics[width=.64\columnwidth]{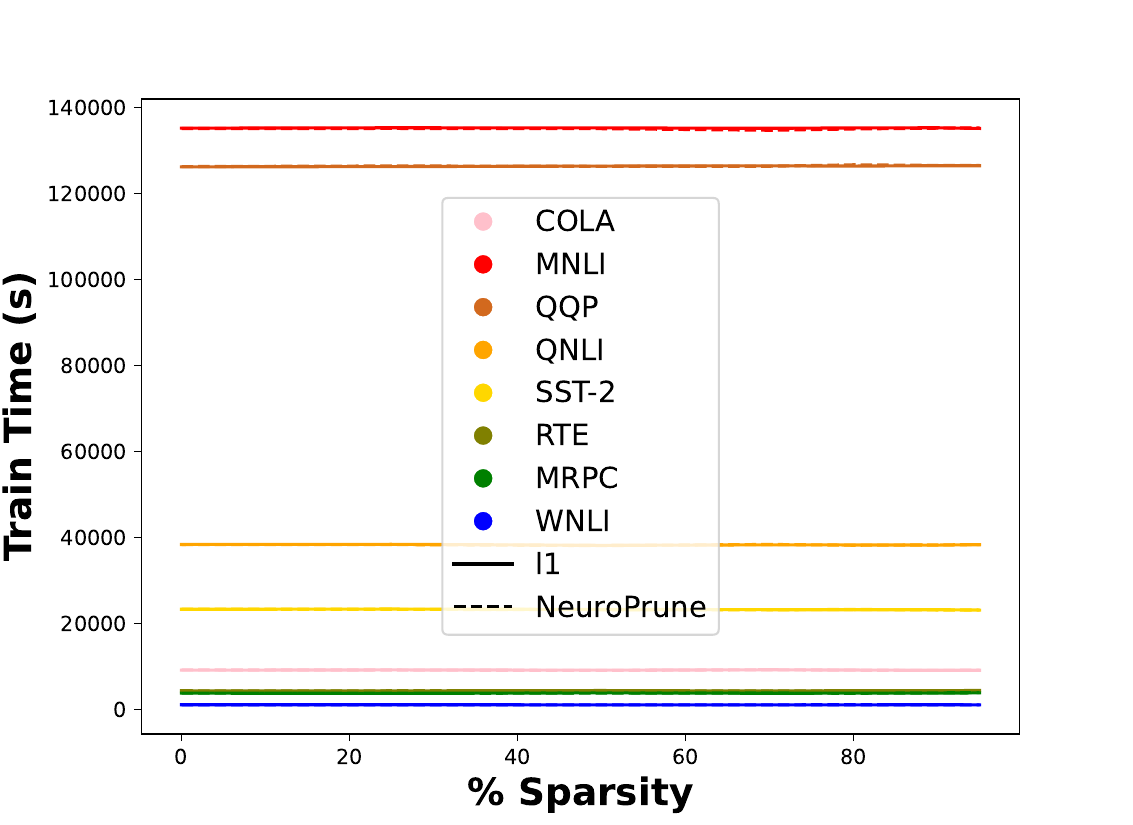}
    \caption{\small{Performance ($1^{\text{st}}$ column), inference time ($2^{\text{nd}}$ column) and train time ($3^{\text{rd}}$ column) for \textsc{NeuroPrune} and $l_1$ on GLUE tasks at different sparsity percentages. The $1^{\text{st}}$ and $2^{\text{nd}}$ rows correspond to T5-large and OPT-1.3b models respectively. In the two rows, we see that \textsc{NeuroPrune} is largely better than $l_1$ sparsity, especially at intermediate sparsities (25-80$\%$), with notable inference time gains and comparable train time. The results are qualitatively similar to those seen with smaller versions of these models in the main paper.
    }}
    \label{fig:sparsity2}
\end{figure*}

Performance with varying sparsity on T5-large and OPT-1.3b are seen in Figure \ref{fig:sparsity2}.







\clearpage
\end{document}